\documentclass[letterpaper, 10 pt, journal, twoside]{IEEEtran}

\IEEEoverridecommandlockouts                                                                 

\usepackage[english]{babel}
\usepackage{times} % assumes new font selection scheme installed
\usepackage{cite}
\usepackage[font=footnotesize]{subcaption}
\usepackage{calrsfs}
\DeclareMathAlphabet{\pazocal}{OMS}{zplm}{m}{n}

\usepackage{cellspace}
\newcolumntype{D}{>{\hfill}N{3}{2}<{\hfill}}

\usepackage{xcolor}
\usepackage[%
	colorlinks = false,
	linkcolor = blue,
	urlcolor  = blue,
	citecolor = blue,
	anchorcolor = blue]{hyperref}

\usepackage{graphicx} % for pdf, bitmapped graphics files
\usepackage{epsfig} % for postscript graphics files
\usepackage{pgfplots}
\usepackage{caption}
\usepackage{subcaption}

\pgfplotsset{compat=1.15}
\graphicspath{{./figure/}}

\usepackage[export]{adjustbox}

\makeatletter
\let\MYcaption\@makecaption
\makeatother

\makeatletter
\let\@makecaption\MYcaption
\makeatother

\usepackage{mathptmx} % assumes new font selection scheme installed
\usepackage{amsmath} % assumes amsmath package installed
\usepackage{amssymb}  % assumes amsmath package installed
\usepackage{nicefrac}
\usepackage{mathtools}
\usepackage{optidef}
\usepackage{leftidx}
\usepackage{bm} % for bold symbols

\def\x#1{\texttt{\expandafter\string\csname#1\endcsname}&\expandafter$\csname#1\endcsname$}

\makeatletter%
\AfterPreamble{%
	\usepackage{hyperref}%
	\let\oldhypertarget\hypertarget%
	\renewcommand{\hypertarget}[2]{%
		\oldhypertarget{#1}{#2}%
		\protected@write\@mainaux{}{%
			\string\expandafter\string\gdef%
			\string\csname\string\detokenize{#1}\string\endcsname{#2}%
		}%
	}%
	\newcommand{\myhyperlink}[1]{%
		\hyperlink{#1}{\csname #1\endcsname}%
	}%
}
\makeatother%

\newcounter{Definition}
\newcommand{\displayDefinitions}[1][]{%
	\stepcounter{Definition}%
	\hypertarget{#1}{\theDefinition}%
}

\newcommand{\refDefinitions}[1][]{%
	\myhyperlink{#1}%
}

\newcounter{Theorem}
\makeatletter
\providecommand{\bigsqcap}{%
	\mathop{%
		\mathpalette\@updown\bigsqcup
	}%
}
\newcommand*{\@updown}[2]{%
	\rotatebox[origin=c]{180}{$\m@th#1#2$}%
}
\makeatother

\DeclareMathOperator*{\maximize}{maximize}

\usepackage{tikz}
\pgfdeclarelayer{foreground}
\pgfsetlayers{background,main,foreground}
\usetikzlibrary{quotes,angles,backgrounds,arrows,automata,shapes,positioning,calc,through,spy}

\tikzset{
    imglabel/.style={
      rectangle,
      inner sep=2pt,
      text=black,
      minimum height=1em,
      text centered,
      fill=white,
      fill opacity=1.0,
      text opacity=1,
      anchor=south west,
    },
  }

\tikzset{
	state/.style={
		rectangle,
		draw=black, very thick,
		minimum height=1.0em,
		text centered,
	},
}

\usepackage{siunitx}
\usepackage{todonotes}

\usepackage{algorithm}
\usepackage[noend]{algpseudocode}

\makeatletter
\def\BState{\State\hskip-\ALG@thistlm}
\makeatother

\usepackage{scalerel}
\usetikzlibrary{svg.path}
\definecolor{orcidlogocol}{HTML}{A6CE39}
\tikzset{
	orcidlogo/.pic={
		\fill[orcidlogocol] 
		svg{M256,128c0,70.7-57.3,128-128,128C57.3,256,0,198.7,0,128C0,57.3,57.3,0,128,0C198.7,0,256,57.3,256,128z};
		\fill[white] svg{M86.3,186.2H70.9V79.1h15.4v48.4V186.2z}
		svg{M108.9,79.1h41.6c39.6,0,57,28.3,57,53.6c0,27.5-21.5,53.6-56.8,53.6h-41.8V79.1z 
		M124.3,172.4h24.5c34.9,0,42.9-26.5,42.9-39.7c0-21.5-13.7-39.7-43.7-39.7h-23.7V172.4z}
		svg{M88.7,56.8c0,5.5-4.5,10.1-10.1,10.1c-5.6,0-10.1-4.6-10.1-10.1c0-5.6,4.5-10.1,10.1-10.1C84.2,46.7,88.7,51.3,88.7,56.8z};
	}
}
\newcommand\orcidicon[1]{\href{https://orcid.org/#1}{\mbox{\scalerel*{
				\begin{tikzpicture}[yscale=-1,transform shape]
				\pic{orcidlogo};
				\end{tikzpicture}
			}{|}}}}

\title{ \LARGE \bf
	Power Line Inspection Tasks with Multi-Aerial Robot Systems via Signal Temporal Logic 
	Specifications
}

\author{Giuseppe Silano$^{1\orcidicon{0000-0002-6816-6002}}$, Tomas 
Baca$^{1\orcidicon{0000-0001-9649-8277}}$, Robert Penicka$^{1\orcidicon{0000-0001-8549-4932}}$, 
Davide Liuzza$^{2\orcidicon{0000-0003-1791-2295}}$, and Martin 
Saska$^{1\orcidicon{0000-0001-7106-3816}}$% <-this % 
	\thanks{Manuscript received: October 15, 2020; Revised December 19, 2020; Accepted February 20, 
	2021.}%
	\thanks{This paper was recommended for publication by  Editor Nancy Amato upon evaluation of 
	the Associate Editor and Reviewers' comments. This work was partially funded by the European 
	Union's Horizon 2020 research and innovation programme AERIAL-CORE under grant agreement no. 
	871479, by CTU grant no. SGS20/174/OHK3/3T/13, and by the Czech Science Foundation (GAČR), 
	within research projects no. 19-22555Y and 20-10280S.}
	\thanks{$^1$Giuseppe Silano, Tomas Baca, Robert Penicka, and Martin Saska are with the Czech 
	Technical University in Prague, Czech Republic, email: {\tt\small 
	\{name.surname\}@fel.cvut.cz.}}
	\thanks{$^2$Davide Liuzza is with the ENEA Fusion and Nuclear Safety Department, Italy, email: 
	{\tt\small davide.liuzza@enea.it.}}
	\thanks{Digital Object Identifier (DOI): see top of this page.}
	
}

\begin{document}
	
%%% INITIAL SUBMISSION RA-L ================================================
	
%\maketitle
%\thispagestyle{empty}
%\pagestyle{empty}

%%% FINAL SUBMISSION RA-L ==================================================

\maketitle
\global\csname @topnum\endcsname 0
\global\csname @botnum\endcsname 0

%%% START SECTION ==========================================================

\begin{abstract}
	
	A framework for computing feasible and constrained trajectories for a fleet of quad-rotors 
	leveraging on Signal Temporal Logic (STL) specifications for power line inspection tasks is 
	proposed in this paper. The planner allows the formulation of complex missions that avoid 
	obstacles and maintain a safe distance between drones while performing the planned mission. An 
	optimization problem is set to generate optimal strategies that satisfy these specifications 
	and also take vehicle constraints into account. Further, an event-triggered replanner is 
	proposed to reply to unforeseen events and external disturbances. An energy minimization term 
	is also considered to implicitly save quad-rotors battery life while carrying out the mission. 
	Numerical simulations in MATLAB and experimental results show the validity and the 
	effectiveness of the proposed approach, and demonstrate its applicability in real-world 
	scenarios.
	
\end{abstract}

%%% END SECTION ============================================================

%%% START SECTION INITIAL SUBMISSION RA-L ==================================

%\begin{keywords}
	
%	Signal temporal logic, multi-robot system, UAV, formal methods-based control, power line 
%	inspection
		
%\end{keywords}

%%% END SECTION ============================================================

%%% START SECTION FINAL SUBMISSION RA-L ====================================

% There is a list of keywords accepted for the letter. They are reported at the link: 
%https://www.ieee-ras.org/publications/ra-l/keywords

%%% END SECTION ============================================================

%%% START SECTION ==========================================================

%\section{Supplementary Material}
%\label{sec:supplementaryMaterial}
%
%Videos with the experiments and numerical simulations can be found
%at~\href{http://mrs.felk.cvut.cz/ral-power-tower-inspection}{mrs.felk.cvut.cz/ral-power-tower-inspection}.

%%% END SECTION ============================================================

%%% START SECTION ==========================================================

\section{Introduction}
\label{sec:introduction}

\IEEEPARstart{O}{ver} the last two decades, global energy demand has increased rapidly due to 
demographic and economic growth. This has created new challenges for electricity supply companies, 
which are constantly looking for new solutions to minimize the frequency of power outages. Power 
failures are particularly critical when the environment and public safety are at risk, e.g., for 
hospitals, sewage treatment plants and telecommunication systems. One of the major causes of a 
power outage is damage to transmission lines, usually due to high winds, storms, or inefficient 
maintenance activities~\cite{EPRI2011TechnicalReport}.

Nowadays, the most common strategy for reducing energy interruptions is to schedule periodic 
inspections using manned helicopters equipped with multiple sensors. Data are captured over 
thousands of kilometers by experienced crews for subsequent processing. There are two major 
drawbacks to this approach: first, flights are dangerous for operators who have to fly close to 
power towers; second, the inspection is extremely time-consuming and expensive (\$1,500 for a 
one-hour flight) and is prone to human error~\cite{Baik2018JIRS, Martinez2018EAAI}.

Multiple solutions have been investigated in the literature for automating this task. Unmanned 
Aerial Vehicles (UAVs) and Rolling On Wire (ROW) robots~\cite{Martinez2018EAAI} have been proposed 
as valuable solutions to replace helicopters within the process. The most promising and the most 
flexible solution is to use UAVs that can perform various levels of inspection depending on the 
wing types and the task of interest~\cite{EPRI2011TechnicalReport}.

%A fixed-wing aircraft can provide a macro-level inspection and can survey power lines over a large 
%area, although the aircraft itself cannot hover. However, a rotary-wing vehicle can exploit its 
%vertical take-off and landing capabilities to carry out a micro-level inspection searching for 
%damage to the mechanical structure and for failures of the electrical components. 

%Indeed, it is extremely important for the whole inspection system to have UAVs with knowledge of 
%the surrounding environment that are capable of planning and executing appropriate actions, 
%starting from (possibly vague) high-level task specifications (e.g., while the UAVs are moving to 
%the target, they must obey the rules for that zone, and must return to the base station in due 
%course). 

However, the use of UAVs to achieve these tasks is particularly challenging, due to the strong 
electromagnetic interference produced by power lines, and the presence of obstacles along the 
line~\cite{Baik2018JIRS}. Accurate task planning is therefore needed to mitigate such issues and to 
accomplish the assigned mission safely. Temporal-Logic (TL) can be of help by providing a powerful 
mathematical tool for the automatic design of feedback control laws that meet complex temporal 
requirements. In particular, Signal Temporal Logic (STL)~\cite{Donze2010FMATS, Maler2004FTMA} can 
be used to describe planning objectives that are more complex than point-to-point planning 
algorithms~\cite{Webb2013ICRA}. This approach leverages on the definition of quantitative 
semantics~\cite{Belta2017Book, Faunekos2009TCS} for TLs to interpret a formula w.r.t. a discretized 
abstraction of the robot motion modelled as a finite transition system. The result is an 
optimization problem with the goal of maximizing a real-valued metric (called \textit{robustness}) 
that denotes how strongly a specification is satisfied or violated. 

%This requirement is extremely important in safety-critical industrial missions such as power line 
%inspection tasks. 

%While alternative solutions allow the problem to be tackled by using an automata-based 
%approach~\cite{Belta2017Book}, optimization-based techniques allow the incorporation of robustness 
%metrics, rather than only a binary answer for the assigned specifications (i.e., satisfied/not 
%satisfied).

%%% END SECTION ============================================================

%%% START SECTION ==========================================================

\subsection{Related works}
\label{sec:relatedWorks}

As detailed in~\cite{Pagnano2013CIRP, Baik2018JIRS}, there are three main challenges for UAVs 
inspecting power lines: (i) \textit{visual servoing} to ensure power line tracking and autonomous 
navigation; (ii) \textit{obstacle detection and avoidance} to prevent possible collisions with the 
towers and obstacles along the path; (iii) \textit{robust control} to provide high stability and 
positioning, hence allowing for close-up inspections. 

Much of the state-of-the-art focuses on the first two problems. Some works~\cite{Chen2019Access, 
Baik2018JIRS} propose new methods for electric tower detection and image segmentation. Others deal 
with the detection of possible mechanical faults or damages to isolation 
material~\cite{Martinez2018EAAI, Pagnano2013CIRP}. In this case, a highly desired feature is a 
control strategy that enables trajectories to be obtained not only for a single vehicle, but 
possibly for a fleet cooperating at the same time, in the same area, while avoiding obstacles and 
possible crashes and respecting the given mission specifications and time bounds. 

%Whether the visual inspection is carried out offline (the drone is teleoperated) or online 
%(waypoints are assigned to the on-board navigation system), the inspection is made with the use of 
%a single drone with a level of autonomy that mainly concerns the vision part. 
%This may improve the overall performance while also reducing costs and enhancing time saving.

Other approaches focus on the endurance of the drone mission as a way to maximize the exploration 
within its battery life. Solutions have been proposed for performing cooperative aerial coverage 
path planning with a multi-UAV system~\cite{Baik2018JIRS, Mansouri2018CEP}. However, these 
problems do not usually take into consideration the dynamics of the drone or physical constraints 
on vehicles. They, therefore, do not offer guarantees on the feasibility of the path and on the 
DOFs of the vehicle motion. Often, the proposed solution is an extension of the well-known 
Traveling Salesman Problem~\cite{Mansouri2018CEP} which, being an NP-hard problem, easily becomes 
unsolvable within a reasonable time when the complexity increases exponentially by the number of 
vehicles and variables.

%To improve the efficiency, the path is planned before the mission begins and the optimal problem 
%is set to cover the given area in minimum time

% COMMENT: This was in the period reported below
%For example, multi-agent planning algorithms have been proposed in~\cite{PantCCT2017, 
%Lindemann2019TCST} for motion control of ground robots and aerial vehicles.

As regards the trajectory planning problem for a multi-robot system, especially for quad-rotors, 
several approaches have been investigated in the literature~\cite{Shoukry2017CDC, Luis2020RAL, 
Honig2018TRO}. Most of the solutions rely on an abstract grid-based representation of the 
environment~\cite{PantCCT2017} or on abstract dynamics of the agents combining a discrete planner 
with a continuous trajectory generator~\cite{Honig2018TRO}. Others propose centralized multi-agent 
path planning methods using relative safe flight corridors to find feasible trajectories for the 
agents~\cite{Park2019IROS}. Although these approaches can compute collision-free trajectories for a 
large number of agents in a short time, they do not offer guarantees that the aircraft will comply 
with the physical constraints or perform the task in a given time window. On the other hand, 
whether the model of quad-rotors is considered~\cite{Luis2020RAL}, the solutions rely on 
information sharing between agents, making them difficult to achieve in presence of electromagnetic 
interference, such as in power line inspection tasks. Moreover, even when the planning algorithms 
demonstrate computational efficiency~\cite{Honig2018TRO}, they do not provide any reference 
regarding velocity and acceleration, leaving the controller to generate these signals.

Many of solutions use Linear Temporal Logic (LTL) as the mission specification language to 
synthesize the optimization problem without considering explicit time bounds on the mission 
objectives~\cite{Shoukry2017CDC}. Other solutions propose the use of STL specifications to describe 
mission requirements without the need to discretize the dynamics or the 
environment~\cite{Raman2014CDC}. Unlike LTL, STL is equipped with qualitative and quantitative 
semantics, meaning that it is not only able to assess whether the system execution meets the 
desired requirements, but also provides a measure of how well the requirements are being met (i.e., 
a \textit{robustness function}). Furthermore, STL semantics takes the absolute time information 
explicitly into account, therefore making it possible to plan when a given task has to be executed 
in the context of the whole mission. 

%However, these approaches are still difficult to apply in real-world applications, because they 
%yield, in their general form, to a NP-hard formulation.
 
%Non-smooth optimization~\cite{PantCCT2017} and stochastic heuristics~\cite{Abbas2013ACM} have been 
%explored to mitigate this problem by providing some guarantees at least for safety properties.

%%% END SECTION ============================================================

%%% START SECTION ==========================================================

\subsection{Contributions}
\label{sec:contributions}

%In particular, the approach allows for a continuous representation of the environment without 
%requiring any \textit{a priori} discretization.

In this paper, we propose a framework for encoding inspection missions for a fleet of quad-rotors 
as STL specifications. Then, using the motion primitives defined in~\cite{Mueller2015TRO}, we 
construct an optimization problem to generate optimal strategies that satisfy the specifications. 
The proposed approach generates feasible dynamic trajectories accounting for the velocity and 
acceleration constraints of the vehicles, avoiding obstacles and maintaining a safe distance 
between drones, while complying with the specifications for the mission. An event-triggered 
replanning strategy is also proposed to account for disturbances and unforeseen events along the 
tracking. The optimization problem is reshaped to compute the feasible path to reconnect the drone 
to the previously computed optimal offline solution. In addition, a minimum energy problem is set 
up to implicitly prevent the quad-rotors from draining the battery while carrying out the mission 
specification successfully.

%In comparison with the approaches described in Sec.~\ref{sec:relatedWorks},

The advantages are twofold: (i) the full expressiveness of the STL formulas allows explicit time 
requirements to be taken into consideration, making the framework easy to reuse and customize for 
applications of interest; (ii) thanks to the motion primitives, the proposed approach can generate 
trajectories in accordance with pre-set velocity and acceleration constraints that can be 
well-tracked by lower-level controllers.

Numerical simulations achieved in MATLAB show the validity of the proposed approach. Various 
scenarios were considered for an evaluation of the trajectory generator performance. A comparison 
between the proposed strategy and an existing stat-of-the-art solution is given at this stage. In 
addition, Gazebo simulations and real-experiments were used to demonstrate the applicability of the 
method in a scenario closer to the real implementation.

\section{Problem Description}
\label{sec:problemDescription}

%The work presented here forms a part of the AERIAL-CORE European project, in which core robotics 
%challenges are tackled using the power line inspection problem as motivation.

The work presented here forms a part of the AERIAL-CORE European project. The \textit{power tower 
inspection} task is considered. A multi-robot system carries out a detailed investigation of power 
equipment, looking for possible faults. The visual examination outputs videos or pictures of 
towers, cable installations, and their surroundings performing a preliminary remote evaluation. The 
aim is to identify components that need to be replaced. %The cameras are mounted in an 
%\textit{eye-in-hand} configuration, i.e., rigidly attached to the body frame. 

We suppose that the UAVs operate in a known environment, represented by a map that also includes 
the position of obstacles and the power tower. Also, that the UAVs are equipped with the necessary 
sensors and software for their own precise localization and state estimation~\cite{Baca2020mrs}. 
%The maps of individual scenarios are obtained from a three-dimensional terrestrial laser scanner. 

%Although it is incomplete, i.e., data are missing in occluded out-of-view locations, this does not 
%affect the planner and it is sufficient for targets localization and mission purposes.

%%% END SECTION ============================================================

%%% START SECTION ==========================================================

\section{Preliminaries}
\label{sec:preliminaries}

Let us consider a continuous-time dynamical system $\pazocal{H}$ and its discrete time version 
$x_{k+1}=f(x_k,u_k)$, where $x_k, x_{k+1} \in X \subset \mathbb{R}^n$ are the current state and the 
next state of the system, respectively, $u \in U \subset \mathbb{R}^m$ is the control input and $f 
\colon X \times U \rightarrow X$ is differentiable in both of the arguments. The initial state is 
denoted by $x_0$ and takes values from some initial set $X_0 \subset \mathbb{R}^n$. Let $T_s \in 
\mathbb{R}_{\geq 0}$ and $T \in \mathbb{R}_{\geq 0}$ be the sampling period and the trajectory 
duration, respectively, so we can write the time interval as the vector $\mathbf{t} = (0, T_s, 
\dots, NT_s)^\top \in \mathbb{R}^{N+1}$, where $NT_s=T$ and $\mathbf{t}_k$, $k \in 
\mathbb{N}_{\geq  0}$, denote the $k$-element of the vector $\mathbf{t}$. Therefore, given an 
initial state $x_0$ and a finite control input sequence $\mathbf{u}=(u_0, \dots, u_{N-1})^\top 
\in \mathbb{R}^N$, a trajectory of the system is the unique sequence of states $\mathbf{x} = (x_0, 
\dots, x_N)^\top \in \mathbb{R}^{N+1}$. Similarly to $\mathbf{t}_k$, with $\mathbf{u}_k$ and 
$\mathbf{x}_k$ we denote the $k$-element of vector $\mathbf{u}$ and $\mathbf{x}$, respectively.

%%% END SECTION ============================================================

%%% START SECTION ==========================================================

\subsection{Signal Temporal Logic}
\label{sec:signalTemporalLogic}

The trajectory generator is designed to satisfy a specification expressed in 
STL~\cite{Donze2010FMATS, Maler2004FTMA}. STL is a logic that allows the succinct and unambiguous 
specification of a wide variety of desired system behaviors over time, such as ``The quad-rotor 
reaches the goal within $10$ time units while always avoiding obstacles''. The semantics of STL are 
defined in~\cite{Maler2004FTMA} and is not reported here for the sake of brevity.

\subsection{Robust Signal Temporal Logic}
\label{sec:robustSignalTemporalLogic}

The presence of a dynamic environment, unforeseen events, and external disturbances can affect the 
closed loop behavior and the satisfaction of the STL formula $\varphi$. For this reason, it is 
convenient to have a maneuverability margin in an attempt to maximize the degree of satisfaction 
with the formula. This can be formally defined and computed using the \textit{robust semantic} of 
temporal logic~\cite{Maler2004FTMA, Faunekos2009TCS, Donze2010FMATS}.

%This degree of satisfaction can be formally defined and computed using the 
%\textit{robust semantic} of temporal logic~\cite{Maler2004FTMA, Faunekos2009TCS, Donze2010FMATS}.

\textit{Definition~\displayDefinitions[STLRobustness]~(Robustness)}: The robustness of an STL 
formula $\varphi$ relative to the system trajectory $\mathbf{x}$ at time $\mathbf{t}_k$ is defined 
via the following recursive formulas
\begin{equation*}
\begin{array}{rll}
%\rho_\top (\mathbf{x}, \mathbf{t}_k) & = & +\infty \\
%
\rho_{p_i} (\mathbf{x}, \mathbf{t}_k) & = & \mu_i (x_{\mathbf{t}_k}), \\ %\forall p_i \in AP \\
\rho_{\neg \varphi} (\mathbf{x}, \mathbf{t}_k) & = & - \rho_\varphi (\mathbf{x}, \mathbf{t}_k), 
\\
\rho_{\varphi_1 \wedge \varphi_2} (\mathbf{x}, \mathbf{t}_k) & = & \min \left(\rho_{\varphi_1} 
(\mathbf{x}, \mathbf{t}_k), \rho_{\varphi_2} (\mathbf{x}, \mathbf{t}_k) \right), \\
\rho_{\square_I \varphi} (\mathbf{x}, \mathbf{t}_k) & = & \min\limits_{\mathbf{t}_k^\prime \in 
	[\mathbf{t}_k + I]} \rho_\varphi (\mathbf{x}, \mathbf{t}_k^\prime), \\
\rho_{\lozenge_I \varphi} (\mathbf{x}, \mathbf{t}_k) & = & \max\limits_{\mathbf{t}_k^\prime \in 
	[\mathbf{t}_k + I]} \rho_\varphi (\mathbf{x}, \mathbf{t}_k^\prime), \\
\rho_{\varphi_1 \pazocal{U} \varphi_2} (\mathbf{x}, \mathbf{t}_k) & = & 
\max\limits_{\mathbf{t}_k^\prime \in [\mathbf{t}_k + I]} \Bigl( \min \left( \rho_{\varphi_2} 
(\mathbf{x}, \mathbf{t}_k^\prime) \right), \\
& &  \hfill \min\limits_{ \mathbf{t}_k^{\prime\prime} \in [\mathbf{t}_k, \mathbf{t}_k^\prime] } 
\left( \rho_{\varphi_1} (\mathbf{x}, \mathbf{t}_k^{\prime \prime} \right)  \Bigr),
\end{array}
\end{equation*}
where $\mathbf{t}_k + I$ is meant here as the Minkowski sum between the scalar $\mathbf{t}_k$ and 
the interval $I$. In the above formulas, $\mu_i(x_{\mathbf{t}_k})$ is a smooth function 
called \textit{predicate} which results true if its value is grater or equal than zero, negative 
otherwise. On example for the robot case could be being inside a target region or being outside an 
obstacle region, with regions described by a certain number of predicates. All the other 
expressions define operators acting on other STL subformulas, thus implicitly describing the 
semantic in a recursive way. Further details can be found in~\cite{Maler2004FTMA, 
Faunekos2009TCS, Donze2010FMATS}. For simplicity, we will write $\rho_\varphi(\mathbf{x})$ instead 
of $\rho_\varphi (\mathbf{x}, 0)$ when $\mathbf{t}_k=0$. Also, we will say that $\mathbf{x}$ 
violates the STL formula $\varphi$ at time $\mathbf{t}_k$ if $\rho_\varphi(\mathbf{x}, 
\mathbf{t}_k) \leq 0$ and that $\mathbf{x}$ satisfies $\varphi$  if $\rho_\varphi(\mathbf{x}, 
\mathbf{t}_k) > 0$. 

Thus, we can compute control inputs $\mathbf{u}$ by maximizing the robustness over the set of 
finite state and input sequences $\mathbf{x}$ and $\mathbf{u}$, respectively. The obtained sequence 
$\mathbf{u}^\star$ is valid if $\rho_\varphi (\mathbf{x}^\star, \mathbf{t}_k)$ is positive, where 
$\mathbf{x}^\star$ and $\mathbf{u}^\star$ obey the dynamical system $\pazocal{H}$. The larger 
$\rho_\varphi (\mathbf{x}^\star, \mathbf{t}_k)$ is, the more robust the behavior of the system is.
 
%Intuitively, in a real world deployment of the drone system, $\mathbf{x}^\star$ can be disturbed 
%and $\rho_\varphi$ might also decrease. Therefore, higher values of such a function offer the 
%possibility of handling higher disturbances before getting negative robustness values, i.e., the 
%STL specification is no longer satisfied. 

\textit{Definition~\displayDefinitions[LSERobustness]~(LSE Robustness)}~\cite{PantCCT2017}: 
Let us consider $c \geq 1$, the smooth approximation of the $m$-array $\max$ and $\min$ is
\begin{equation*}
\begin{split}
&\max(\rho_{\varphi_1}, \dots, \rho_{\varphi_m}) \approx \frac{1}{c} \log \left( 
\sum_{i=1}^m  e^{c \rho_{\varphi_i}} \right), 
\\
&\min(\rho_{\varphi_1}, \dots, \rho_{\varphi_m}) \approx -\frac{1}{c} \log \left( \sum_{i=1}^m 
e^{-c \rho_{\varphi_i}} \right). 
\end{split}
\end{equation*}
This log-sum-exponential (LSE) approximation is smooth, and an analytical form of its gradient 
exists. This robustness approximation approaches the true robustness values given according to 
Def.~\refDefinitions[STLRobustness] as $c \rightarrow \infty$. The larger $c$ is, the 
greater the accuracy of the approximation is.

%%% END SECTION ============================================================

%%% START SECTION ==========================================================

\section{Problem Formulation}
\label{sec:problemFormulation}

%In this section, we show how to generate trajectories for a fleet of $q$ quad-rotors starting 
%from mission specifications expressed as an STL formula. The motion planner is the result of an 
%optimization problem that outputs a global path for the vehicles. The use of motion primitives 
%accounts for the vehicle constraints and ensures the retrieval of feasible paths from the 
%optimization. These paths are used as a reference by the trajectory tracking controller that 
%performs the inspection. Figure~\ref{fig:overallSystem} describes the overall system architecture.

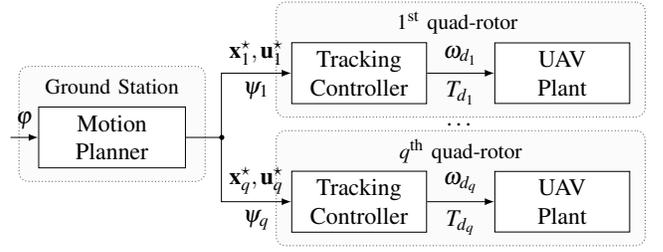
\begin{figure}[t]
	\begin{center}
		\scalebox{0.9}{
		\begin{tikzpicture}
		
		%%%%%%%%%%%%%%%%%%%%% Motion Planner
		% Ground station block
		\node (MultiRobots-Box) at (0,0.2) [fill=gray!3,rounded corners, draw=black!70, densely 
		dotted, minimum height=1.7cm, minimum width=2.75cm]{}; 
		
		% Motion Planner
		\node (MotionPlanner) at (0,0) [text centered, fill=white, draw, rectangle, minimum 
		width=1.5cm, text width=5.5em]{Motion\\Planner};
		
		% Links
		\draw[-latex] ($(MotionPlanner) - (1.5,0)$) -- node[above]{$\varphi$} (MotionPlanner);
		
		% Multi-robots box
		\draw[fill=black] ($ (MotionPlanner.east) + (0.5,0) $) arc(-180:180:0.03);
		\node (GroundStation) at (0,0.75) [text centered]{\small Ground Station};
		
		%%%%%%%%%%%%%%%%%%%
		% Drone 1
		\node (Drone-Box1) at (5.15,1.15) [fill=gray!3,rounded corners, draw=black!70, densely 
		dotted, minimum height=1.7cm, minimum width=5.45cm]{}; 
		\node (TrackingController1) at (3.65,0.95) [fill=white, draw, rectangle, text centered, 
		text width=5em]{Tracking\\Controller};
		\node (UAVPlant1) at (6.65,0.95) [fill=white, draw, rectangle, text centered, text 
		width=5em]{UAV\\Plant};
		\node (Drone-Box1-Text) at (5.1,1.7) [text centered]{\small $1$\textsuperscript{st} 
			quad-rotor};	
		
		% Drone 1 - Links
		\draw[-latex] (TrackingController1) -- node[above]{$\omega_{d_1}$} node[below]{$T_{d_1}$} 
		(UAVPlant1);
		\draw[-latex] (MotionPlanner.east) -- ($ (MotionPlanner.east) + (0.525,0) $) 
		-- ($ (MotionPlanner.east) + (0.525,0.95) $) -- node[above]{$\mathbf{x}^\star_1, 
		\mathbf{u}^\star_1$} node[below]{$\psi_1$} (TrackingController1.west);
		
		%%%%%%%%%%%%%%%%%%%
%		% Drone 2
%		\node (Drone-Box2) at (5.15,0.2) [fill=gray!3,rounded corners, draw=black!70, densely 
%		dotted, minimum height=1.7cm, minimum width=5.45cm]{};
%		\node (TrackingController2) at (3.65,0) [fill=white, draw, rectangle, text centered, text 
%		width=5em]{Tracking\\Controller};
%		\node (UAVPlant2) at (6.65,0) [fill=white, draw, rectangle, text centered, text 
%		width=5em]{UAV\\Plant};
%		\node (Drone-Box2-Text) at (5.15,0.75) [text centered]{\small $2$\textsuperscript{nd} 
%			quad-rotor};		
%		
%		% Drone 2 - Links
%		\draw[-latex] (TrackingController2) -- node[above]{$\omega_{d_2}$} node[below]{$T_{d_2}$} 
%		(UAVPlant2);	 
%		\draw[-latex] ($ (MotionPlanner.east) + (0.525,0) $) -- 
%		node[above]{$\mathbf{x}^\star_2, \mathbf{u}^\star_2$} node[below]{$\psi_2$} 
%		(TrackingController2.west);
		
		%%%%%%%%%%%%%%%%%%%
		% Drone J
		\node (Drone-BoxN) at (5.15,-0.75) [fill=gray!3,rounded corners, draw=black!70, densely 
		dotted, minimum height=1.7cm, minimum width=5.45cm]{};
		% Dots
		\node (Dots2) at (5.15,0.2) [text centered]{\dots};
		\node (TrackingControllerN) at (3.65,-0.95) [fill=white, draw, rectangle, text centered, 
		text width=5em]{Tracking\\Controller};
		\node (UAVPlantN) at (6.65,-0.95) [fill=white, draw, rectangle, text centered, text 
		width=5em]{UAV\\Plant}; 	
		\node (Drone-BoxN-Text) at (5.15,-0.2) [text centered]{\small $q$\textsuperscript{th} 
			quad-rotor};
		
		% Drone J - Links
		\draw[-latex] (TrackingControllerN) -- node[above]{$\omega_{d_q}$} node[below]{$T_{d_q}$} 
		(UAVPlantN);
		\draw[-latex] ($ (MotionPlanner.east) + (0.525,0) $) -- ($ 	
		(MotionPlanner.east) + (0.525,-0.95) $) -- node[above]{$\mathbf{x}^\star_q, 
		\mathbf{u}^\star_q$} node[below]{$\psi_q$} (TrackingControllerN.west);
		
		\end{tikzpicture}
		}
		%\vspace*{-5mm}
		\caption{Control scheme. The motion planner generates the trajectories $\left( 
		\mathbf{x}_i^\star, \mathbf{u}_i^\star \right)$ and the heading angles $\psi_i$, with 
		$i=\{1, \dots, q\}$, for the $q$ quad-rotors by using the STL mission specifications 
		$\varphi$. A tracking controller supplies the desired angular velocities $\omega_{d_i}$ and 
		thrust $T_{d_i}$ commands for the UAVs.}
		\label{fig:overallSystem}
	\end{center}
\vspace{-1.5em}
\end{figure}

In this section, we show how to generate trajectories for a fleet of $q$ quad-rotors starting 
from mission specifications $\varphi$. The motion planner is the result of an optimization problem 
that outputs a global feasible path for the vehicles accounting for their constraints. These 
paths are used as a reference by the trajectory tracking controller that performs the inspection. 
Figure~\ref{fig:overallSystem} describes the overall system architecture.

\subsection{Motion planner}
\label{sec:motionPlanner}

The use of an STL robust semantic allows to synthesize the motion planner, i.e., finding a control 
sequence for the $q$ quad-rotors that satisfies a given STL formula $\varphi$. Such a problem is 
casted for each quad-rotor as an optimization problem over the control $\mathbf{u} = (u_0, \dots, 
u_{N-1})^\top$ and state $\mathbf{x} = (x_0, \dots, x_{N-1})^\top$ sequences as follows
\begin{equation}\label{eq:optimizationProblem}
\begin{split}
&\maximize_{\mathbf{u},\,\mathbf{x}} \;\; {\rho_\varphi(\mathbf{x})} \\
&\quad \;\;\; \text{s.t.}~\quad\;\; \mathbf{x}_{k+1}= f(\mathbf{x}_k, \mathbf{u}_k), \forall 
k=\{0, 1, \dots, N-1\} \\
\end{split},
\end{equation}
where $\mathbf{x}_0 = x_0$. Note that, in order to make this paper more readable, 
in~\eqref{eq:optimizationProblem} we provided the optimization problem for each quad-rotor, 
considering them decoupled. However, in the case of coupling among some of them, such as for a 
minimum distance to be always kept, problem~\eqref{eq:optimizationProblem} can be analogously 
written taking into account the state and control sequences of all the involved vehicles as 
decision variables, as well as their dynamics. The coupling constraint can be embedded in the STL 
formula used in the objective function. 

%where $\rho_\varphi(\mathbf{x}) \in \mathbb{R}$ and $\mathbf{x}_0 = x_0$. If the optimal control 
%sequence $\mathbf{u}^\star \in \mathbb{R}^N$ generates an output signal $\mathbf{x}^\star \in 
%\mathbb{R}^N$, then we find a trajectory that ensures satisfaction of the specification. 

As detailed in Def.~\refDefinitions[STLRobustness], $\rho_\varphi$ uses non-differentiable 
functions $\max$ and $\min$. Therefore, the robustness of the STL formula $\varphi$ is itself 
non-differentiable as a function of the trajectory $\mathbf{x}$ and the control inputs 
$\mathbf{u}$. While mixed-integer programming solvers~\cite{Raman2014CDC}, non-smooth 
optimizers, or stochastic heuristics~\cite{Abbas2013ACM} can be used to find a solution for this 
problem, the problem is NP-hard, and these approaches could fail with the increase of the number of 
variables. However, as shown in~\cite{PantCCT2017}, a good approach for mitigating computational 
complexity is to adopt a smooth approximation $\tilde{\rho}_\varphi$ of the robust function 
$\rho_\varphi$. One of the possible choices is LSE robustness 
(Def.~\refDefinitions[LSERobustness]). In this case, the resulting optimization problem is still 
non-convex, but smooth optimization techniques, such as sequential quadratic programming, can be 
used to find a local maximum. In this paper, such an approach is adopted to compute the robustness 
value by using the smooth operator defined in~\cite{PantCCT2017}.

To come up with a trajectory that satisfies the vehicle constraints, the motion	primitives defined 
in~\cite{Mueller2015TRO} have been considered. The method allows for obtaining rapid generation 
and feasibility verification of motion primitives for quad-rotors. Let us define the state 
$\mathbf{x}$ and control $\mathbf{u}$ sequences as $\mathbf{x}_k=(\mathbf{p}^{(1)}_k, 
\mathbf{v}^{(1)}_k, \mathbf{p}^{(2)}_k, \mathbf{v}^{(2)}_k, \mathbf{p}^{(3)}_k, 
\mathbf{v}^{(3)}_k)^\top$ and $\mathbf{u}_k=(\mathbf{a}^{(1)}_k, \mathbf{a}^{(2)}_k, 
\mathbf{a}^{(3)}_k)^\top$, where $\mathbf{p}_k^{(j)}$, $\mathbf{v}_k^{(j)}$, and 
$\mathbf{a}^{(j)}_k$, with $j=\{1,2,3\}$, represent the vehicle's position, velocity, and 
acceleration at time instant $k$ along the $j$-axis of the inertial frame, respectively. The 
optimization problem~\eqref{eq:optimizationProblem} can be reformulated approximating the 
translational dynamics of the quad-rotor separately along each $j$-axis with the splines 
$\mathbf{S}^{(j)} ( \mathbf{p}_k^{(j)}, \mathbf{v}_k^{(j)}, \mathbf{a}_k^{(j)} ) = 
(\mathbf{p}_{k+1}^{(j)}, \mathbf{v}_{k+1}^{(j)}, \mathbf{a}_{k+1}^{(j)})^\top$ defined as
\begin{equation}\label{eq:splines}
\mathbf{S}^{(j)} = 
\begin{pmatrix}
\frac{\alpha}{120} \mathbf{t}_k^5 + \frac{\beta}{24} \mathbf{t}_k^4 + \frac{\gamma}{6} 
\mathbf{t}_k^3 + \mathbf{a}_k^{(j)} \mathbf{t}_k^2 + \mathbf{v}_k^{(j)} \mathbf{t}_k + 
\mathbf{p}_k^{(j)} \\
\frac{\alpha}{24} \mathbf{t}_k^4 + \frac{\beta}{6} \mathbf{t}_k^3 + \frac{\gamma}{2} 
\mathbf{t}_k^2 + \mathbf{a}_k^{(j)} \mathbf{t}_k + \mathbf{v}_k^{(j)} \\
\frac{\alpha}{6} \mathbf{t}_k^3 + \frac{\beta}{2} \mathbf{t}_k^2 + \gamma \, \mathbf{t}_k + 
\mathbf{a}_k^{(j)}
\end{pmatrix},
\end{equation}
where $\mathbf{p}^{(j)}_0=p^{(j)}_0$, $\mathbf{v}^{(j)}_0=v^{(j)}_0$, and 
$\mathbf{a}^{(j)}_0=a^{(j)}_0$, while parameters $\alpha$, $\beta$, and $\gamma$  that can be tuned 
to achieve a desired motion fixing a combination of position, velocity, and acceleration at the 
start and end points~\cite[Appx.~A]{Mueller2015TRO}. Such an approach ensures compliance with 
safety requirements and intrinsically embeds the gravity 
compensation~\cite[Sec.~III]{Mueller2015TRO}. Thus, the accelerations $\mathbf{a}^{(j)}$ are meant 
as the variations w.r.t. the vertical equilibrium position.

%Although more advanced planning techniques are available in the 
%literature~\cite{Mellinger2011ICRA}, most impose a rigid structure on the end state, e.g., 
%allowing a fixed or varying maneuver duration or specifying the goal state with convex 
%inequalities. Moreover, some of these techniques are characterized by a huge computational demand. 
%This may increase the inherently NP-hard complexity issues of TL optimization problems and move to 
%an unfeasible solution. The benefits of using this approach are twofold: first, the framework 
%allows the generation of trajectories for an arbitrary maneuverer duration and state constraints; 
%second, the required computational burden is of the order of a million motion primitives per 
%second, which can be generated and tested on an ordinary laptop~\cite{Mueller2015TRO}.

Thus, the problem~\eqref{eq:optimizationProblem} can be reformulated replacing 
$\rho_\varphi(\mathbf{x})$ with its smooth version $\tilde{\rho}_\varphi(\mathbf{x})$ considering 
for the mathematical formulation of the trajectory generator $\mathbf{S}^{(j)}$. Moreover, 
exploiting the decoupling of the drone dynamics into three orthogonal 
axes~\cite[Sec.~III-C]{Mueller2015TRO}, the original optimization 
problem~\eqref{eq:optimizationProblem} can be split into three independent problems for each 
$j$-axis, as follows
\begin{equation}\label{eq:optimizationProblemMotionPrimitives}
\begin{split}
&\maximize_{\mathbf{p}^{(j)}, \mathbf{v}^{(j)},\,\mathbf{a}^{(j)}} \;\;
{\tilde{\rho}_\varphi (\mathbf{p}^{(j)}, \mathbf{v}^{(j)} )} \\
&\quad \,\;\, \text{s.t.}~\quad\;\;\; \lvert \mathbf{v}^{(j)}_k \rvert \leq 
\mathbf{v}^{(j)}_\mathrm{max}, \lvert \mathbf{a}^{(j)}_k \vert  \leq \mathbf{a}^{(j)}_\mathrm{max}, 
\\
&\,\;\;\;\;\, \qquad \quad\;\;\; \text{eq.}~\eqref{eq:splines}, \forall k=\{0,1, \dots, N-1\}
\end{split},
\end{equation}
where $\mathbf{v}^{(j)}_\mathrm{max}$ and $\mathbf{a}^{(j)}_\mathrm{max}$ are the desired maximum 
values of velocity and acceleration along the motion, respectively. The higher $N$ is, the bigger 
the number of DOFs is. Consequently, the computational burden for solving the optimization problem 
increases. However, smaller values of $N$ restrict the DOFs of the motion planner, thus potentially 
providing a trajectory that does not satisfy the STL specification. While the acceleration 
$\mathbf{a}^{(j)}$ is bounded in norm in the optimization 
problem~\eqref{eq:optimizationProblemMotionPrimitives}, the bound on the jerk is implicitly 
accounted by the chosen motion primitives in~\cite{Mueller2015TRO}. %The spline based STL 
%optimization problem provided in~\eqref{eq:optimizationProblemMotionPrimitives} concurs in the 
%core of the proposed framework, and forms a part of the contribution of our paper together 
%with the event-triggered and energy-aware planners (Secs.~\ref{sec:eventTriggeredReplanner 
%and~\ref{sec:minimumEnergyProblem}).} %Further details on the values of the parameters 
%employed in the optimization problem are provided in Sec.~\ref{sec:numericalResults}, 

%It is worth noticing that the acceleration $\mathbf{a}^{(j)}$ is bounded in norm in the 
%optimization problem~\eqref{eq:optimizationProblemMotionPrimitives}. Therefore, the possibility to 
%obtain from such problem time instants in which the acceleration is negative is, in principle, 
%possible since it is not explicitly excluded with a dedicated constraint. However, according 
%to the problem formulation~\eqref{eq:optimizationProblemMotionPrimitives}, the optimal solution 
%does not imply, for ``forward'' motion, an acceleration that turns negative for some time. This is 
%why reverse acceleration is never obtained as solution of the problem (see 
%Figs.~\ref{fig:graphConstraintsPowerTower} and~\ref{fig:energyAccelerationRobustness}). Whereas, 
%the bound on the jerk is implicitly accounted by the chosen motion primitives 
%in~\cite{Mueller2015TRO}.

%%% END SECTION ============================================================

%%% START SECTION ==========================================================

\subsection{Event-triggered replanner}
\label{sec:eventTriggeredReplanner}

As explained in the previous section, the adoption of motion primitives allows to obtain 
feasible solutions for the quad-rotors dynamics. It may be the case that, due to unexpected large 
disturbances at runtime, a significant mismatch between the planned trajectory and the quad-rotor 
state can be experienced. To cope with such an issue, here we introduce an online event-based 
replanner. %Event-triggered approaches~\cite{MaityACC2018} are recently gaining attention, also for 
%multi-agent system case~\cite{YuTCNS2019}.

Specifically, in our case we consider to obtain data only at certain discrete time instances 
denoted by $\bar{\mathbf{t}}$. Let $T_e \in \mathbb{R}_{\geq 0}$ and $T_g \in \mathbb{R}_{\geq 0}$ 
be the event-triggering period (multiple of the sampling period $T_s$) and the ``topic'' 
waypoint period (a low-rate sequence of the state $\mathbf{x}$, with $T_g >\!\!> T_s$), 
respectively, so we can write the discrete time instances $\bar{\mathbf{t}}$ and $\hat{\mathbf{t}}$ 
as the vectors $\bar{\mathbf{t}} = (0, T_e, \dots, LT_e)^\top \in \mathbb{R}^{L+1}$ and 
$\hat{\mathbf{t}} = (0, T_g, \dots, GT_g)^\top \in \mathbb{R}^{G+1}$, where $LT_e \subseteq 
T$ and $GT_g \subset T$. The term $\bar{\mathbf{t}}_l$, $l \in \mathbb{N}_{\geq 0}$, denotes the 
$l$-element of the vector $\bar{\mathbf{t}} \subseteq \mathbf{t}$, while $\hat{\mathbf{t}}_g$, $g 
\in \mathbb{N}_{\geq 0}$, denotes the $g$-element of the vector $\hat{\mathbf{t}} \subset 
\mathbf{t}$. 

We also denote with $\tilde{\mathbf{p}}$ the runtime trajectory position of the drone. Notice that 
such trajectory could be different from the optimal one $\mathbf{p}^\star$ due to disturbances 
acting at runtime.

At each time instant, say $\bar{\mathbf{t}}_l \in \bar{\mathbf{t}}$ the condition $\lvert 
\tilde{\mathbf{p}}_l - \mathbf{p}_l \rvert > \eta$ is evaluated, with $\eta > 0$ a design parameter 
triggering threshold. If such condition results true, then a trigger is generated and the actual 
drone position is communicated to the ground station. The latter performs an optimal replanning 
operation over the time interval $\{\bar{\mathbf{t}}_l, \hat{\mathbf{t}}_{g+1}\}$, where  
$\hat{\mathbf{t}}_{g+1}$ is the time associated with the next topic position $\mathbf{p}_{g+1}$. In 
this way, the feasible path between the triggering position $\mathbf{p}_l$ and the next position 
$\mathbf{p}_{g+1}$ is computed.

%The event-triggered mechanism is based on a minimum triggering interval $\Delta \mathbf{\bar{t}} = 
%\mathbf{\bar{t}}_l - \mathbf{\bar{t}}_{l-1}$, for each $l \geq 1$, and on a norm bound 
%$\mathbf{e}_p$ defined as the difference between the value of the current position $\mathbf{p}_k$ 
%and the value of the position at the last triggering instance $\mathbf{p}_l$. The presence of a 
%triggering interval allows to relax the need of a continuous communication between the drone and 
%the planner, saving energy and bandwidth to transmit continuous measurements. In contrast to 
%continuous feedback replanner, the event-triggered requires less state measurements and ensures 
%good performance if the triggering interval $\Delta \mathbf{\bar{t}}$ is sufficiently small. This 
%planner relies on a event generator that decides it at the instances $\mathbf{\bar{t}}$ the 
%quad-rotors state measurements need to be sent to the planner\footnote{For sake of simplicity, the 
%event-triggering period $T_e$ was chosen as half of the sampling period $T_s$.}.

%%% END SECTION ============================================================

%%% START SECTION ==========================================================

\subsection{Energy-aware planner}
\label{sec:minimumEnergyProblem}

The synthesized motion planner problem in Sec.~\ref{sec:motionPlanner} can be modified to ensure 
that the quad-rotors also save their battery charge while carrying out their mission successfully. 
The objective is to generate a trajectory for the $q$ quad-rotors that also takes into account the 
energy requirement of the vehicles.

Let us define the decision variables $\bm{\epsilon}_k =  (\bm{\epsilon}^{(1)}_k, 
\bm{\epsilon}^{(2)}_k, \bm{\epsilon}^{(3)}_k)^\top$, where $\bm{\epsilon}^{(j)}_k$, with 
$j=\{1,2,3\}$, represents the bound on the square norm of the quad-rotor acceleration (i.e., 
the control input) as a proxy of the energy at time instant $k$ along the $j$-axis of the inertial 
frame. As discussed in~\cite{Honig2018TRO, Luis2020RAL}, the optimal trajectory that deals with 
energy minimization can be obtained by minimizing the positive semi-definitive quadratic form 
$\bm{\epsilon}^\top_k \mathbf{Q} \bm{\epsilon}_k$, where $\mathbf{Q} \in \mathbb{R}^{3N \times 3N}$ 
such that for all $\bm{\epsilon}_k \in \mathbb{R}^{3N}$ we have that $\bm{\epsilon}_k^\top 
\mathbf{Q} \bm{\epsilon}_k \geq 0$. Thus, the optimization 
problem~\eqref{eq:optimizationProblemMotionPrimitives} can be reformulated by adding a new term to 
the cost function and bounding the system energy $\lVert {\mathbf{a}^{(j)}}^\top \mathbf{a}^{(j)} 
\rVert^2$. Namely we write:
\begin{equation}\label{eq:energeryAwareMotionPlanner}
\begin{split}
&\maximize_{\mathbf{p}^{(j)}, \mathbf{v}^{(j)},\,\mathbf{a}^{(j)},\,\bm{\epsilon}^{(j)}} \;\;
{\tilde{\rho}_\varphi (\mathbf{p}^{(j)}, \mathbf{v}^{(j)} ) - {\bm{\epsilon}^{(j)}}^\top \mathbf{Q} 
\bm{\epsilon}^{(j)}} \\
&\quad\quad \;\, \text{s.t.}~\qquad\;\, \lvert \mathbf{v}^{(j)}_k \rvert \leq 
\mathbf{v}^{(j)}_\mathrm{max}, \lvert \mathbf{a}^{(j)}_k \vert  \leq 
\mathbf{a}^{(j)}_\mathrm{max}, \\
&\quad\;\;\;\;\, \qquad \qquad\; \lVert {\mathbf{a}^{(j)}_k}^\top \mathbf{a}^{(j)}_k \rVert^2 \leq 
{\bm{\epsilon}^{(j)}_k}^\top \bm{\epsilon}_k^{(j)}, \bm{\epsilon}_k^{(j)} \geq 0, \\
&\quad\;\;\;\;\, \qquad \qquad\;\, \text{eq.}~\eqref{eq:splines}, \forall k=\{0, 1, \dots, N-1\} \\
\end{split}.
\end{equation}
The optimization problem both incorporates the satisfaction of the STL formula $\varphi$ and 
the energy saving to prevent that the drones run out of battery while performing the mission, at 
the expense of a reduction of the robustness $\rho_\varphi(\mathbf{x})$. 

%%% END SECTION ============================================================

%%% START SECTION ==========================================================
	
\subsection{Control architecture}
\label{sec:controlArchitecture}

The control architecture is reported in Fig.~\ref{fig:overallSystem}. Starting from mission and 
vehicle constraints, the \textit{Motion Planner} solves the optimization 
problem~\eqref{eq:optimizationProblemMotionPrimitives} supplying the trajectories $\left( 
\mathbf{x}^\star, \mathbf{u}^\star \right)$ and the heading angles $\psi$ (provided as a constant 
reference for each target) for the $q$ quad-rotors. The trajectory generation is run 
\textit{one-shot}, i.e., once at time $\mathbf{t}_k = 0$, and the result is used as reference by 
the tracking controller. 

%The obstacles, the positions of the power tower and the workspace dimensions are also considered 
%at this stage. The trajectory generation is run \textit{one-shot}, i.e., once at time 
%$\mathbf{t}_k = 0$, and the result is used as reference by the drone trajectory tracking 
%controller. 

%Starting from~\cite{Baca2020mrs}, we designed the control architecture (see 
%Fig.~\ref{fig:lowLevelController}). This is divided into two parts: the high-level layer, i.e., 
%\textit{Reference Tracker} and \textit{Reference Controller}, which generates the desired angular 
%velocities $\omega_d$ and thrust $T_d$ command signals, starting from the optimization outputs and 
%the low-level layer, i.e., \textit{Attitude Rate Controller}, which combines a non-linear $SO(3)$ 
%controller and onboard sensor data, to compute the propeller speed $\tau_d$.

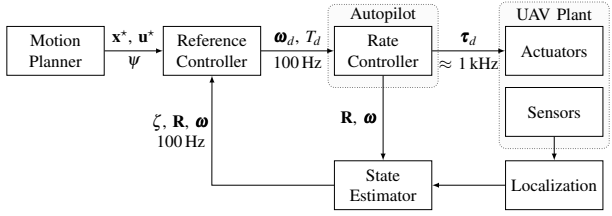
\begin{figure}
	\centering
	\scalebox{0.65}{
		\begin{tikzpicture}
		
		%%%%% COLORED REGIONS
		\node (Region1) at (6,0.1) [fill=gray!3, rounded corners, draw=black!70, densely dotted, 
		minimum height=1.7cm, minimum width=2.25cm]{}; % Rate Controller
		\node (Region2) at (9.5,-0.5) [fill=gray!3, rounded corners, draw=black!70, densely dotted, 
		minimum height=3.0cm, minimum width=2.25cm]{}; % Actuators
		%\node (Region3) at (-1,0.1) [fill=gray!3, rounded corners, draw=black!70, densely dotted, 
		%minimum height=1.7cm, minimum width=2.9cm]{}; % Motion Planner
		%\node (Region4) at (2.5,0.1) [fill=gray!3, rounded corners, draw=black!70, densely dotted, 
		%minimum height=1.7cm, minimum width=2.25cm]{}; % Reference Generator
		%\node (Region5) at (6,-2.9) [fill=gray!3, rounded corners, draw=black!70, densely dotted, 
		%minimum height=1.7cm, minimum width=2.25cm]{}; % State Estimator
		%\node (Region6) at (9.5,-2.9) [fill=gray!3, rounded corners, draw=black!70, densely 
		%dotted, minimum height=1.7cm, minimum width=2.25cm]{}; % Localization
		\node (UAVPlant-Text) at (9.5,0.75) [text centered]{UAV Plant};
		\node (RateController-Text) at (6,0.725) [text centered]{Autopilot};
		%\node (MotionPlanner-Text) at (-1,0.725) [text centered]{Onboard};
		%\node (ReferenceController-Text) at (2.5,0.725) [text centered]{Onboard};
		%\node (StateEstimator-Text) at (6,-3.475) [text centered]{Onboard};
		%\node (Localization-Text) at (9.5,-3.475) [text centered]{Onboard};
		
		%%%%% NODES
		\node (MotionPlanner) at (-0.7,0) [draw, rectangle, minimum width=1cm, minimum 
		height=1cm, text centered, text width=5em, fill=white]{Motion\\Planner};
		\node (ReferenceController) at (2.5,0) [draw, rectangle, minimum width=1cm, minimum 
		height=1cm, text centered, text width=5em, fill=white]{Reference\\Controller};
		\node (RateController) at (6,0) [draw, rectangle, minimum width=1cm, minimum height=1cm, 
		text centered, text width=5em, fill=white]{Rate\\Controller};
		\node (Actuators) at (9.5,0) [draw, rectangle, minimum width=1cm, minimum height=1cm, text 
		centered, text width=5em, fill=white]{Actuators};
		\node (Sensors) at (9.5,-1.25) [draw, rectangle, minimum width=1cm, minimum height=1cm, 
		text centered, text width=5em, fill=white]{Sensors};
		\node (Localization) at (9.5,-2.75) [draw, rectangle, minimum width=1cm, minimum 
		height=1cm, text centered, text width=5em, fill=white]{Localization};
		\node (StateEstimator) at (6,-2.75) [draw, rectangle, minimum width=1cm, minimum 
		height=1cm, text centered, text width=5em, fill=white]{State\\Estimator};
		
		%%%%% PATHS
		\draw[-latex] (MotionPlanner) -- node[above]{$\mathbf{x}^\star$, 
		$\mathbf{u}^\star$} node[below]{$\psi$} (ReferenceController);
		\draw[-latex] (ReferenceController) -- node[above]{$\bm{\omega}_d$, $T_d$} 
		node[below]{\SI{100}{\hertz}} (RateController);
		\draw[-latex] (RateController) -- node[above]{$\bm{\tau}_d$} node[below]{$\approx$ 
		\SI{1}{\kilo\hertz}} (Actuators);
		\draw[-latex] (Sensors.south) -- (Localization.north);
		\draw[-latex] (Localization.west) -- (StateEstimator.east);
		\draw[latex-] (StateEstimator.north) -- node[left]{$\mathbf{R}$, $\bm{\omega}$} 
		(RateController.south);
		\draw[-latex] (StateEstimator.west) -- ($ (StateEstimator.west) - (2.5,0) $) -- node[text 
		width=7em]{$\zeta$, $\mathbf{R}$, $\bm{\omega}$\\ \;\SI{100}{\hertz}} 
		(ReferenceController.south);
		
		\end{tikzpicture}
	}
	% \vspace*{-6mm}
	\caption{The control architecture. The \textit{Motion Planner} supplies the trajectory 
	$\left( \mathbf{x}^\star, \mathbf{u}^\star \right)$ and the heading angle $\psi$ to the 
	\textit{Reference Controller}, which outputs the thrust $T_d$ and angular velocities 
	$\bm{\omega}_d$ for the embedded \textit{Rate Controller}. A \textit{State Estimator} 
	provides the UAV translation and rotation ($\zeta$, $\mathbf{R}$).}
\label{fig:lowLevelController}
\end{figure}

In Fig.~\ref{fig:lowLevelController} the designed control architecture based on~\cite{Baca2020mrs} 
is reported. This is divided into two parts: the high-level layer, i.e., \textit{Reference 
Controller}, which generates the desired angular velocities $\bm{\omega}_d$ and thrust $T_d$ 
command signals, by using the optimization outputs and the low-level layer, i.e., \textit{Rate 
Controller}, which computes the propellers speed $\bm{\tau}_d$.

\section{Numerical Results}
\label{sec:numericalResults}

To prove the validity and the effectiveness of the proposed approach, we carried out numerical 
simulations in MATLAB, extracting the needed STL specifications from the problem description (see 
Sec.~\ref{sec:problemDescription}). At this stage, the vehicle dynamics and the trajectory 
tracking controller are not considered. The Gazebo robotics simulator was used in the second 
step to numerically verify the feasibility of the trajectories, exploiting the advantages of 
Software-in-the-loop simulations~\cite{Silano2019SMC}. In particular, Gazebo simulations 
were used to reduce the probability of failures and to obtain a qualitative analysis of the system 
behavior. The framework was coded using the 2019b release of MATLAB, with the optimization problem 
formulated using CASADI library and NLP as solver. All simulations were performed on a laptop with 
an i7-8565U processor ($\SI{1.80}{\giga\hertz}$) and $32$GB of RAM running on Ubuntu 18.04. Videos 
with the experiments and numerical simulations in MATLAB and Gazebo are available 
at~\url{http://mrs.felk.cvut.cz/ral-power-tower-inspection}.

%The maximum number of iterations was set to $2000$, with acceptable tolerance for the optimization 
%problem of $\num{1.0e-4}$.

%Experimental results are instead given later in a dedicated section.

%%% END SECTION ============================================================

%%% START SECTION ==========================================================

\subsection{Power tower inspection}
\label{sec:powerTowerInspection}

\begin{figure}
	\begin{center}
		%\hspace*{-12mm}
		% left - bottom - right - top
		\adjincludegraphics[trim={{.075\width} {.0\height} {0.0\width}			
			{.0\height}},clip,scale=0.205] 
		{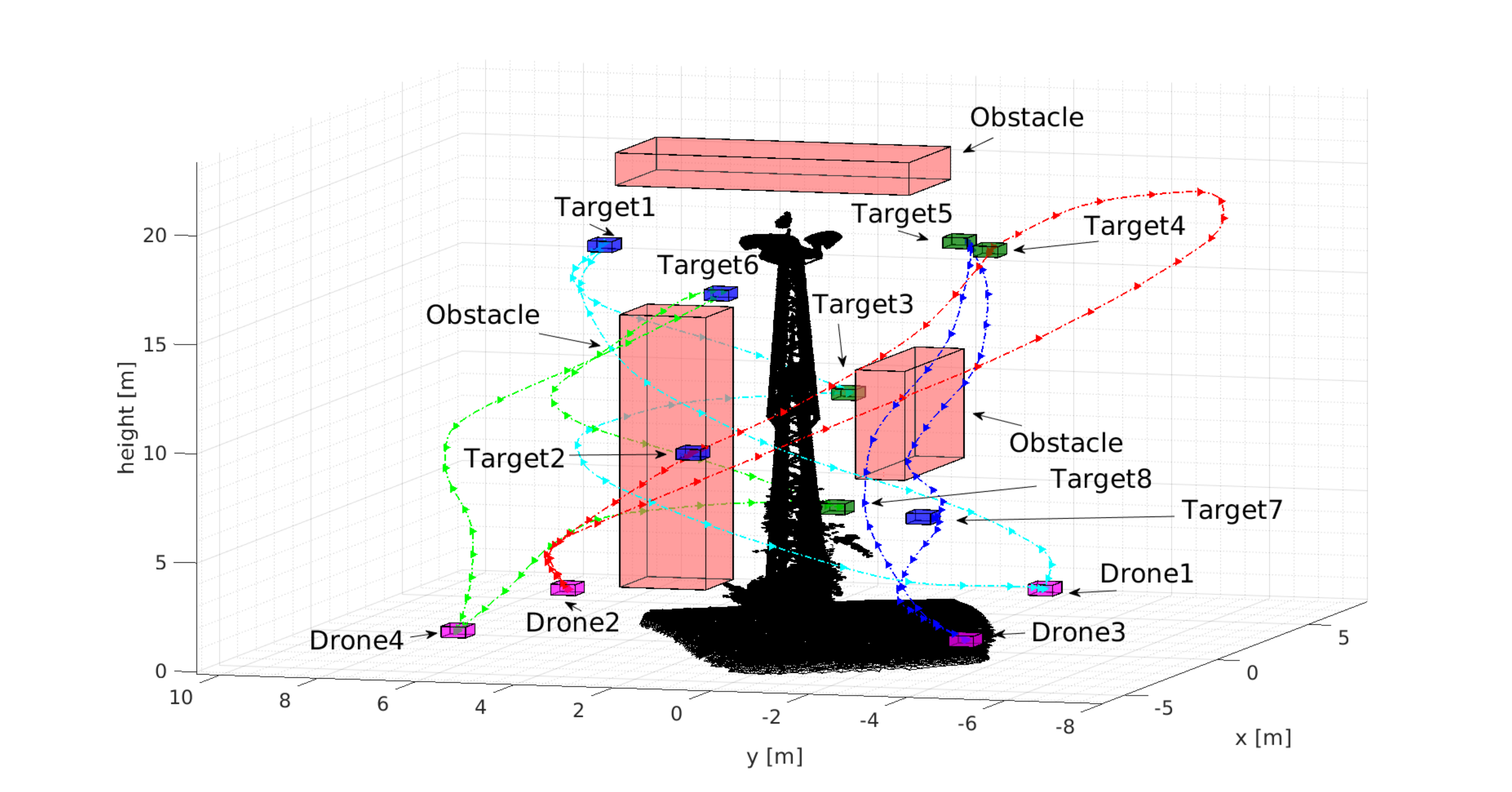}
		\vspace*{-9mm}
	\end{center}
	\caption{\textit{Power tower inspection} scenario when considering four drones and eight 	
	target regions.}
	\label{fig:powerTowerInspectionMoreTargetRegions}
	\vspace{-1em}
\end{figure}

The task objective is to reach target regions (i.e., interesting areas to inspect) within the time 
interval $[0, \nicefrac{2T}{3}]$ while staying within the workspace area ($[14 \si{\meter} \times 
18 \si{\meter} \times 23 \si{\meter}]$), avoiding possible collisions with the power tower and the 
obstacles along the path, and maintaining a safe distance ($\delta_\mathrm{min}$) between drones. 
The mission ends with the drones returning to the initial position within the time interval 
$[\nicefrac{2T}{3}, T]$. When the drones reach the target regions, they start collecting images and 
videos by simulating a data acquisition process. To minimize the time required for 
inspection, the target regions are clustered to find a balance among the number of vehicles 
available for the inspection. However, advanced clustering algorithms may be used accounting for 
the drones positions and the distance between targets. For ease of experimentation, we considered 
only two drones and four target regions, but this does not imply a loss of generality of the 
approach. A numerical simulation was also carried out in MATLAB to show the feasibility of 
the problem as the number of drones and target regions increases (see 
Fig.~\ref{fig:powerTowerInspectionMoreTargetRegions}). The task objective can be encoded with STL 
specifications as follows
\begin{equation}\label{eq:longRangeInspection}
\resizebox{0.95\hsize}{!}{$
	\begin{split}
	\prescript{i}{h}{\varphi_\mathrm{dis}} &= \square_{[0, T]} \left(\lVert 
	\prescript{i}{}{\mathbf{p}} - \prescript{h}{}{\mathbf{p}} \rVert  \geq \delta_\mathrm{min} 
	\right), \\% \quad
	%
	%\prescript{i}{}{\varphi_\mathrm{safe}} = \square_{[0, T]} \Bigl( \lVert 
	%\prescript{i}{}{\mathbf{p}} - \mathbf{p}_\mathrm{obs} \rVert  > 0 \Bigr) , \\
	%
	\varphi_\mathrm{lri} &= \mathop{\wedge}_{{k, i}, {\,k \neq i}}^q \Bigl( 
	\prescript{k}{i}{\varphi_\mathrm{dis}} \wedge \prescript{k}{}{\varphi_\mathrm{safe}} \wedge 
	\prescript{k}{}{\varphi_\mathrm{ws}} \Bigr) \bigwedge \lozenge_{[0,T]} \biggl( \Bigl( 
	\mathop{\wedge}^{\nicefrac{q}{w}}_{k=1} \prescript{k}{}{\varphi_\mathrm{tr1}} \wedge 
	\prescript{k}{}{\varphi_\mathrm{tr3}} \\ 
	&~\bigwedge \mathop{\wedge}_{k=\nicefrac{q}{w}+1}^{q} \prescript{k}{}{\varphi_\mathrm{tr2}} 
	\wedge \prescript{k}{}{\varphi_\mathrm{tr4}} \Bigr) \pazocal{U}_{[\nicefrac{2T}{3} ,T]} \left( 
	\mathop{\wedge}_{k=1}^q \prescript{k}{}{\varphi_\mathrm{home}} \right) \biggr), 
	\end{split}
	$}
\end{equation}
where $\prescript{i}{h}{\varphi_\mathrm{dis}}$ represents the safety distance requirement 
between the $i$-th drone and the $h$-th drone, $\varphi_\mathrm{ws}$, $\varphi_\mathrm{safe}$, and 
$\varphi_\mathrm{home}$ indicate the workspace, safety (i.e., avoiding collisions with the power 
tower and with obstacles along the path), and starting point specifications, respectively, $w$ is 
the number of clusters, while $\varphi_\mathrm{tr1}$, $\varphi_\mathrm{tr2}$, 
$\varphi_\mathrm{tr3}$, and $\varphi_\mathrm{tr4}$ are the target regions. 
%
%\begin{table}
%	\centering
%	\begin{adjustbox}{max width=\columnwidth}
%		\begin{tabular}{|l|c|c|c|c|c|}
%			\hline
%			& \textbf{Sym.} & \textbf{Value-ML} & \textbf{Value-PT} & \textbf{Value-EA} & 
%			\textbf{Unit} 
%			\\
%			\hline
%			LSE scaling factor & $c$ & 10 & 5 & 5 & \si{1} \\
%			Drone safe distance & $\delta_\mathrm{min}$ & 3 & 3 & 3 & \si{\meter} \\
%			Sampling time & $T_s$ & $0.05$ & $0.05$ & $0.05$ & \si{\second} \\
%			Maximum velocity & $\mathbf{v}^{(j)}_\mathrm{max}$ & $3$ & $3$ & $3$ & 
%			\si{\meter\per\second} \\
%			Maximum acceleration & $\mathbf{a}^{(j)}_\mathrm{max}$ & $2.5$ & $3$ & $3$ & 
%			\si{\meter\per\square\second} \\
%			Trajectory period & $T$ & $110$ & $60$ & $110$ &\si{\second} \\
%			\hline
%		\end{tabular}
%	\end{adjustbox}
%	\caption{Optimization problem parameter values for the \textit{power tower inspection} and 
%		\textit{multi-line} scenarios.}
%	\label{tab:tableParamters}
%\end{table}
%
\begin{table}
	\centering
	\begin{adjustbox}{max width=\columnwidth}
		\begin{tabular}{|l|c|c|c|c|}
			\hline
			& \textbf{Sym.} & \textbf{Value-PT} & \textbf{Value-EA} & \textbf{Unit} 
			\\
			\hline
			LSE scaling factor & $c$ & 5 & 5 & \si{1} \\
			Drone safe distance & $\delta_\mathrm{min}$ & 3 & 3 & \si{\meter} \\
			Sampling period & $T_s$ & $0.05$ & $0.05$ & \si{\second} \\
			Maximum velocity & $\mathbf{v}^{(j)}_\mathrm{max}$ & $3$ & $3$ & 
			\si{\meter\per\second} \\
			Maximum acceleration & $\mathbf{a}^{(j)}_\mathrm{max}$ & $3$ & $3$ & 
			\si{\meter\per\square\second} \\
			Trajectory duration & $T$ & $60$ & $110$ &\si{\second} \\
			\hline
		\end{tabular}
	\end{adjustbox}
	\caption{Optimization problem parameter values.}
	\label{tab:tableParamters}
\end{table}

%The mission requirements define a state space region of the system $\pazocal{H}$ according to the 
%definition of an STL formula~\cite{Maler2004FTMA}.

\begin{figure}
	\begin{center}
		%\hspace*{-12mm}
		% left - bottom - right - top
		\adjincludegraphics[trim={{.075\width} {.0\height} {0.0\width}			
			{.0\height}},clip,scale=0.205]{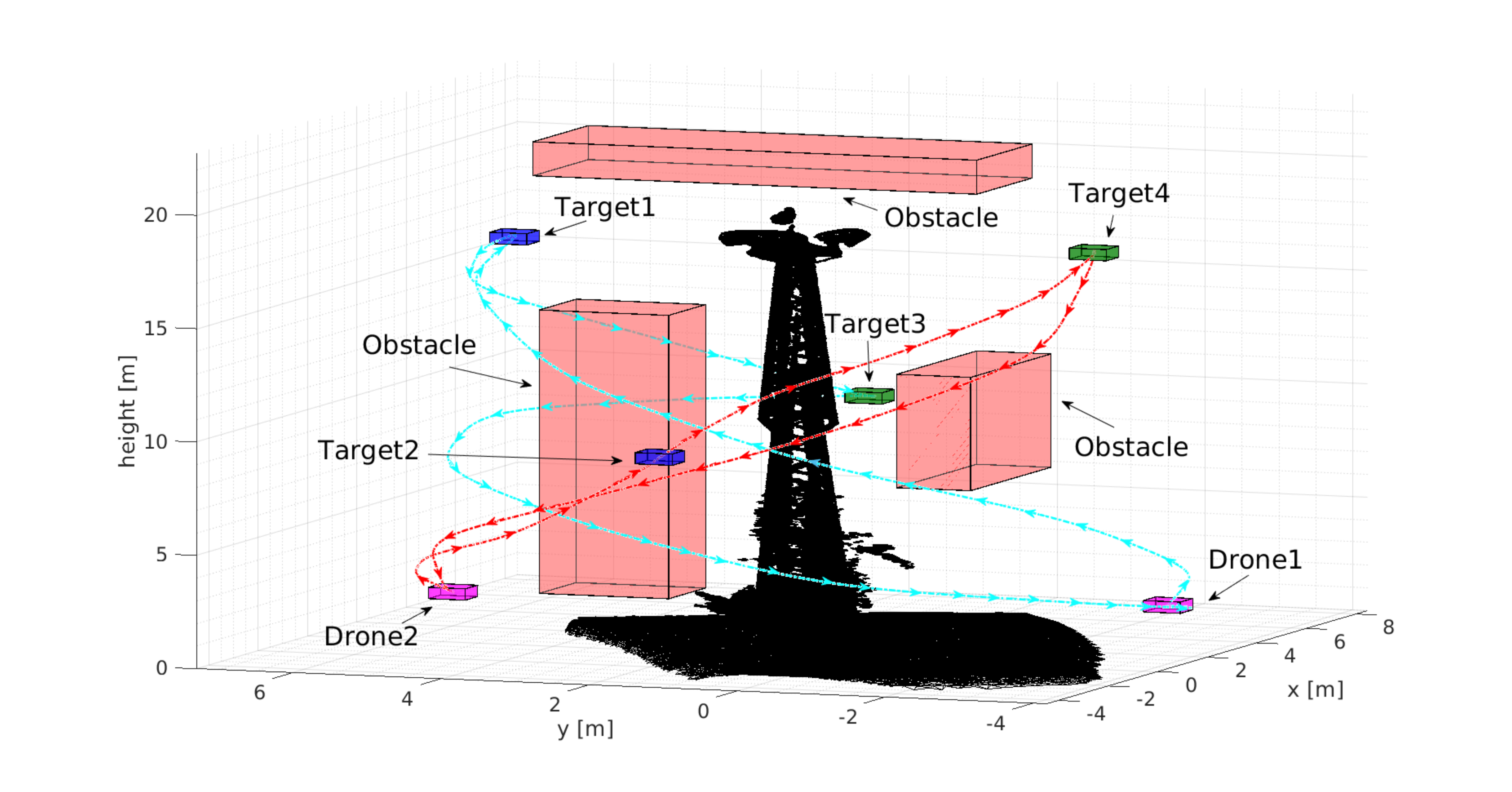}
		\vspace*{-9mm}
	\end{center}
	\caption{\textit{Power tower inspection} scenario. Target regions are represented in blue and 
	green and reflect the navigation order, respectively. Obstacles are depicted in red, while 
	the starting points are in magenta.}
	\label{fig:powerTowerInspection}
	\vspace{-1em}
\end{figure}

The scenario is depicted in Figs.~\ref{fig:powerTowerInspection} 
and~\ref{fig:powerTowerInspectionMoreTargetRegions} along with the obtained trajectories, obstacles 
and target regions (both the obstacles and the target regions are modeled as polyhedra). The 3D map 
was obtained from a three-dimensional terrestrial laser scan of the environment and contains an 
observation tower with a camera and some lights placed on top. These were chosen as regions of 
interest for the inspection. The tower is $\SI{20}{\meter}$ in height with a radius of 
$\SI{3}{\meter}$. Table~\ref{tab:tableParamters} reports the optimization problem and the parameter 
values for the drone considered in this paper, namely DJI F450. The optimization took 
$\SI{21}{\second}$ to solve in the scenario with four target regions and two drones, and 
$\SI{43}{\second}$ in the scenario with eight target regions and four drones. 

Gazebo simulations were performed to qualitatively and quantitatively analyze the time advantages 
deriving from the use of multiple drones to perform the inspection of a power tower w.r.t. using 
only one drone. The scenario reported in Fig.~\ref{fig:powerTowerInspectionMoreTargetRegions} was 
used as a testbed showing that the time required for the inspection took $\SI{60}{\second}$ in the 
multi-UAV case and $\SI{255}{\second}$ for the case of a single quad-rotor.

\subsection{Energy-aware and event-triggered planner}
\label{sec:energyAwareEventTriggeredReplanner}

%The mission requirements (i.e., maximum velocity and acceleration and LSE scaling factor, see 
%Tab.~\ref{tab:tableParamters}) remained unchanged in both cases. 

\begin{figure}
	\begin{center}
		%\hspace*{-12mm}
		% left - bottom - right - top
		\adjincludegraphics[trim={{.075\width} {.0\height} {0.0\width}			
			{.0\height}},clip,scale=0.205] {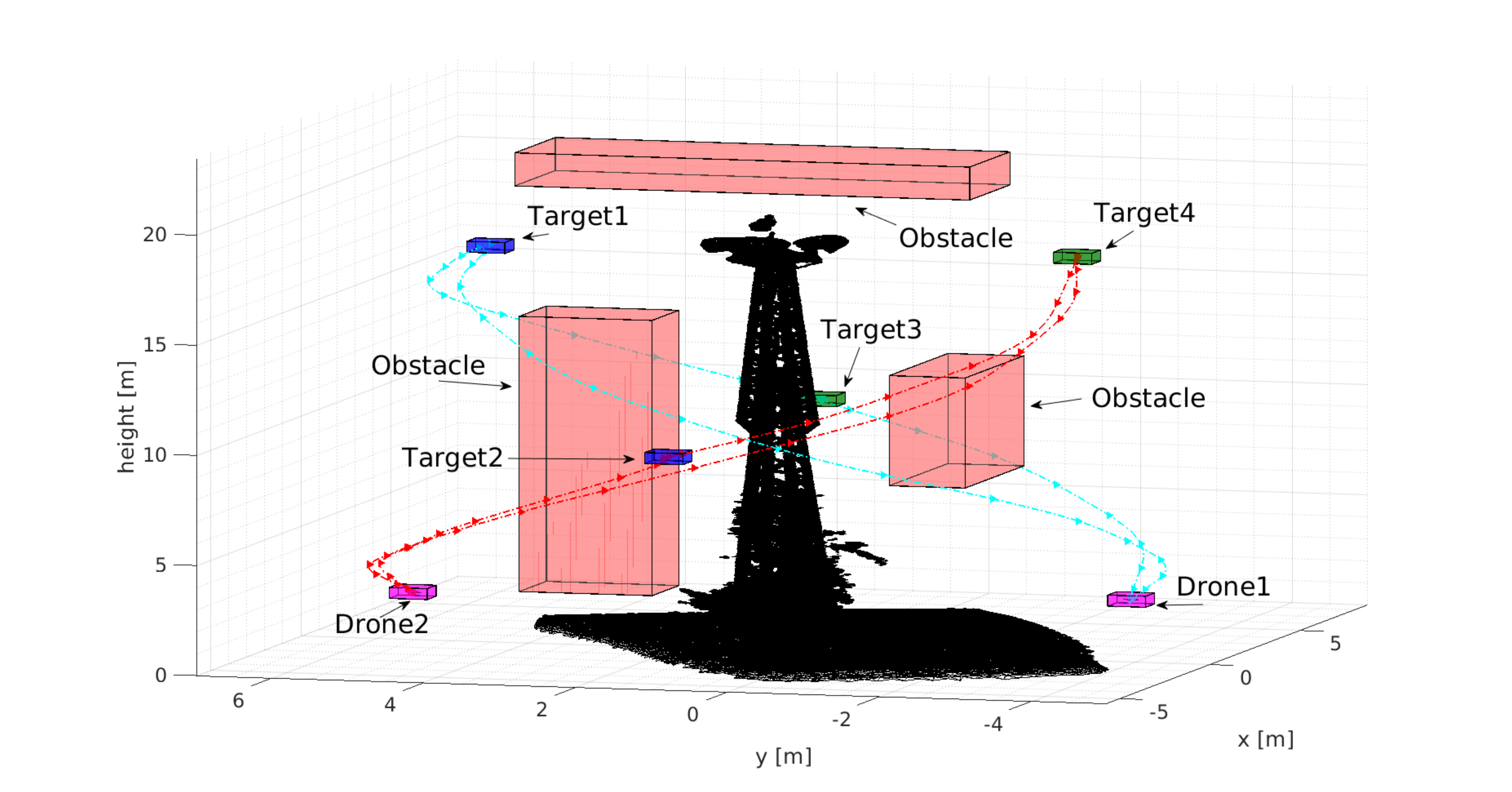}
		\vspace*{-9mm}
	\end{center}
	\caption{\textit{Power tower inspection} scenario considering the energy-aware motion planner.}
	\label{fig:powerTowerInspectionEnergyMinimization}
\end{figure}

\begin{figure}
	\begin{center}
		%\hspace*{-12mm}
		% left - bottom - right - top
		\adjincludegraphics[trim={{.075\width} {.0\height} {0.0\width}			
			{.0\height}},clip,scale=0.205] 
		{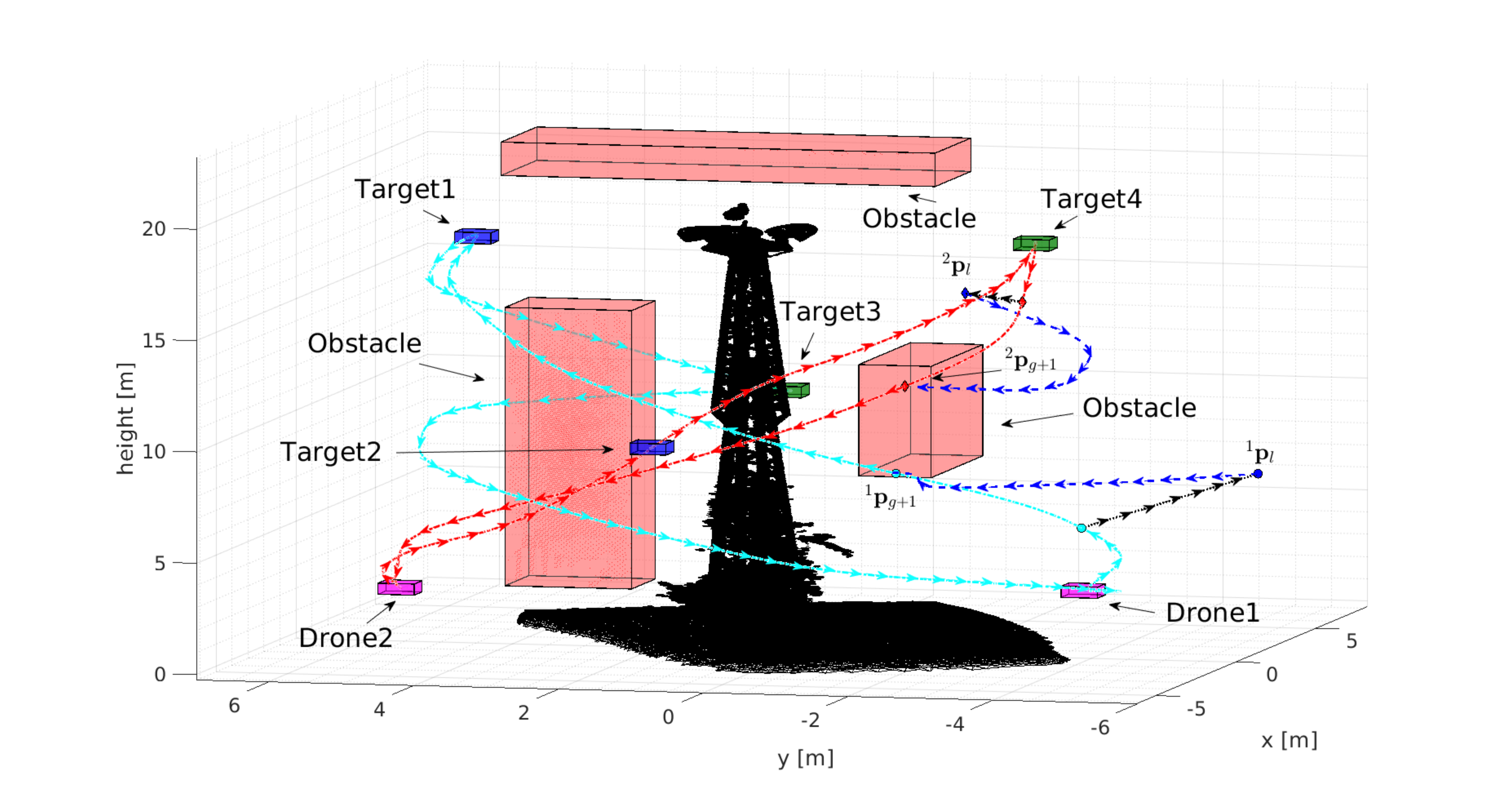}
		\vspace*{-9mm}
	\end{center}
	\caption{\textit{Power tower inspection} scenario in case of unexpected large disturbances 
	events at runtime. Arrows represent the drones' path along the mission. In black the deviation 
	from the original, in blue the new path computed by the replanner.}
	\label{fig:powerTowerEventTriggeredReplanner}
	\vspace{-1em}
\end{figure}

\begin{figure}
	\vspace{-0.50em}
	\begin{center}
	\hspace{-0.725cm}
	\begin{subfigure}{0.45\columnwidth}
		\scalebox{0.52}{
			\begin{tikzpicture}
			\begin{axis}[%
			width=2.8119in,%
			height=1.8183in,%
			at={(0.758in,0.481in)},%
			scale only axis,%
			xmin=0,%
			xmax=60,%
			ymax=20,%
			ymin=-10,%
			xmajorgrids,%
			ymajorgrids,%
			ylabel style={yshift=-0.415cm}, %shifting the y line text
			%xlabel={Time [\si{\second}]},%
			ylabel={[\si{\meter}]},%
			axis background/.style={fill=white},%
			legend style={at={(0.50,0.60)},anchor=north,legend cell 
				align=left,draw=none,legend columns=-1,align=left,draw=white!15!black}
			]
			\addplot [color=blue, dotted, line width=0.75pt] 
			file{matlabPlots/Power_Tower_Inspection/position_UAV42_x.txt};%
			\addplot [color=red, dashed, line width=0.75pt] 
			file{matlabPlots/Power_Tower_Inspection/position_UAV42_y.txt};%
			\addplot [color=green, solid, line width=0.75pt] 
			file{matlabPlots/Power_Tower_Inspection/position_UAV42_z.txt};%
			\draw [thick, dashed] (160, -10) -- (160, 20);
			\draw [thick, dashed] (320, -10) -- (320, 20);
			\fill[blue!50,nearly transparent] (0,20) -- (20,20) -- (20,-10) -- (0,-10) -- 
			cycle;
			\fill[red!50,nearly transparent] (20,20) -- (40,20) -- (40,-10) -- (20,-10) -- 
			cycle;
			\fill[green!50,nearly transparent] (40,20) -- (60,20) -- (60,-10) -- (40,-10) 
			-- cycle;
			\legend{$p^{(1)}$, $p^{(2)}$, $p^{(3)}$};%
			\node [draw,fill=white] at (rel axis cs: 0.2,0.85) {\shortstack[l]{drone1}};
			\end{axis}
			\end{tikzpicture}
			}
		\end{subfigure}
		\hspace{0.25cm}
		\begin{subfigure}{0.45\columnwidth}
		\scalebox{0.52}{
			\begin{tikzpicture}
			\begin{axis}[%
			width=2.8119in,%
			height=1.8183in,%
			at={(0.758in,0.481in)},%
			scale only axis,%
			xmin=0,%
			xmax=60,%
			ymax=20,%
			ymin=-10,%
			xmajorgrids,%
			ymajorgrids,%
			ylabel style={yshift=-0.415cm}, %shifting the y line text
			%xlabel={Time [\si{\second}]},%
			ylabel={[\si{\meter}]},%
			axis background/.style={fill=white},%
			legend style={at={(0.45,0.625)},anchor=north,legend cell 
				align=left,draw=none,legend columns=-1,align=left,draw=white!15!black}
			]
			\addplot [color=blue, dotted, line width=0.75pt] 				
			file{matlabPlots/Power_Tower_Inspection/position_UAV53_x.txt};%
			\addplot [color=red, dashed, line width=0.75pt] 
			file{matlabPlots/Power_Tower_Inspection/position_UAV53_y.txt};%
			\addplot [color=green, solid, line width=0.75pt] 
			file{matlabPlots/Power_Tower_Inspection/position_UAV53_z.txt};%
			\draw [thick, dashed] (160, -10) -- (160, 20);
			\draw [thick, dashed] (320, -10) -- (320, 20);
			\fill[blue!50,nearly transparent] (0,20) -- (20,20) -- (20,-10) -- (0,-10) -- 
			cycle;
			\fill[red!50,nearly transparent] (20,20) -- (40,20) -- (40,-10) -- (20,-10) -- 
			cycle;
			\fill[green!50,nearly transparent] (40,20) -- (60,20) -- (60,-10) -- (40,-10) 
			-- cycle;
			\legend{$p^{(1)}$, $p^{(2)}$, $p^{(3)}$};%
			\node [draw,fill=white] at (rel axis cs: 0.8,0.85) {\shortstack[l]{drone2}};
			\end{axis}
			\end{tikzpicture}
			}
		\end{subfigure}
		\\
		\vspace{0.05cm}
		\hspace{-0.95cm}
		\begin{subfigure}{0.45\columnwidth}
			\centering
			\scalebox{0.52}{
				\begin{tikzpicture}
				\begin{axis}[%
				width=2.8119in,%
				height=1.8183in,%
				at={(0.758in,0.481in)},%
				scale only axis,%
				xmin=0,%
				xmax=60,%
				ymax=4,%
				ymin=-4,%
				xmajorgrids,%
				ymajorgrids,%
				extra y ticks={-3,3},
				extra y tick labels={$-\mathbf{v}^{(j)}_\mathrm{max}$, 
				$\mathbf{v}^{(j)}_\mathrm{max}$},
				extra y tick style={ticklabel style={xshift=0pt}}, % if they crash with the 
				%default ticks
				ylabel style={yshift=-0.775cm}, %shifting the y line text
				%xlabel={Time [\si{\second}]},%
				ylabel={[\si{\meter\per\second}]},%
				axis background/.style={fill=white},%
				legend style={at={(0.50,0.85)},anchor=north,legend cell 
					align=left,draw=none,legend columns=-1,align=left,draw=white!15!black}
				]
				\addplot [color=blue, dotted, line width=0.75pt] 
				file{matlabPlots/Power_Tower_Inspection/velocity_UAV42_x.txt};%
				\addplot [color=red, dashed, line width=0.75pt] 
				file{matlabPlots/Power_Tower_Inspection/velocity_UAV42_y.txt};%
				\addplot [color=green, solid, line width=0.75pt] 
				file{matlabPlots/Power_Tower_Inspection/velocity_UAV42_z.txt};%
				\draw [thick, dashed] (0, 3) -- (450, 3);
				\draw [thick, dashed] (0, -3) -- (450, -3);
				\draw [thick, dashed] (160, -4) -- (160, 4);
				\draw [thick, dashed] (320, -4) -- (320, 4);
				\fill[blue!50,nearly transparent] (0,3) -- (20,3) -- (20,-3) -- (0,-3) -- 
				cycle;
				\fill[red!50,nearly transparent] (20,3) -- (40,3) -- (40,-3) -- (20,-3) -- 
				cycle;
				\fill[green!50,nearly transparent] (40,3) -- (60,3) -- (60,-3) -- (40,-3) 
				-- cycle;
				\legend{$v^{(1)}$, $v^{(2)}$, $v^{(3)}$};%
				\end{axis}
				\end{tikzpicture}
			}
		\end{subfigure}
		\hspace{0.25cm}
		\begin{subfigure}{0.45\columnwidth}
			\centering
			\scalebox{0.52}{
				\begin{tikzpicture}
				\begin{axis}[%
				width=2.8119in,%
				height=1.8183in,%
				at={(0.758in,0.481in)},%
				scale only axis,%
				xmin=0,%
				xmax=60,%
				ymax=4,%
				ymin=-4,%
				xmajorgrids,%
				ymajorgrids,%
				extra y ticks={-3,3},
				extra y tick labels={$-\mathbf{v}^{(j)}_\mathrm{max}$, 
				$\mathbf{v}^{(j)}_\mathrm{max}$},
				extra y tick style={ticklabel style={xshift=0pt}}, % if they crash with the 
				%default ticks
				ylabel style={yshift=-0.775cm}, %shifting the y line text
				%xlabel={Time [\si{\second}]},%
				ylabel={[\si{\meter\per\second}]},%
				axis background/.style={fill=white},%
				legend style={at={(0.50,0.825)},anchor=north,legend cell 
					align=left,draw=none,legend columns=-1,align=left,draw=white!15!black}
				]
				\addplot [color=blue, dotted, line width=0.75pt] 
				file{matlabPlots/Power_Tower_Inspection/velocity_UAV53_x.txt};%
				\addplot [color=red, dashed, line width=0.75pt] 
				file{matlabPlots/Power_Tower_Inspection/velocity_UAV53_y.txt};%
				\addplot [color=green, solid, line width=0.75pt] 
				file{matlabPlots/Power_Tower_Inspection/velocity_UAV53_z.txt};%
				\draw [thick, dashed] (0, 3) -- (450, 3);
				\draw [thick, dashed] (0, -3) -- (450, -3);
				\draw [thick, dashed] (160, -4) -- (160, 4);
				\draw [thick, dashed] (320, -4) -- (320, 4);
				\fill[blue!50,nearly transparent] (0,3) -- (20,3) -- (20,-3) -- (0,-3) -- cycle;
				\fill[red!50,nearly transparent] (20,3) -- (40,3) -- (40,-3) -- (20,-3) -- 
				cycle;
				\fill[green!50,nearly transparent] (40,3) -- (60,3) -- (60,-3) -- (40,-3) -- 
				cycle;
				\legend{$v^{(1)}$, $v^{(2)}$, $v^{(3)}$};%
				\end{axis}
				\end{tikzpicture}
			}
		\end{subfigure}
		\\
		\vspace{0.05cm}
		\hspace{-0.95cm}
		\begin{subfigure}{0.45\columnwidth}
			\centering
			\scalebox{0.52}{
				\begin{tikzpicture}
				\begin{axis}[%
				width=2.8119in,%
				height=1.8183in,%
				at={(0.758in,0.481in)},%
				scale only axis,%
				xmin=0,%
				xmax=60,%
				ymax=4,%
				ymin=-4,%
				xmajorgrids,%
				ymajorgrids,%
				extra y ticks={-3,3},
				extra y tick labels={$-\mathbf{a}^{(j)}_\mathrm{max}$, 
				$\mathbf{a}^{(j)}_\mathrm{max}$},
				extra y tick style={ticklabel style={xshift=0pt}}, % if they crash with the 
				%default ticks
				%extra x ticks={20,40},
				%extra x tick labels={$\varphi_\mathrm{tr1} \text{,} \, \varphi_\mathrm{tr2}$, 
				%	$\varphi_\mathrm{tr3} \text{,} \, \varphi_\mathrm{tr4}$},
				%extra x tick style={ticklabel style={yshift=-10pt}}, % if they crash with the 
				%default ticks
				ylabel style={yshift=-0.775cm}, %shifting the y line text
				%xlabel={Time [\si{\second}]},%
				ylabel={[\si{\meter\per\square\second}]},%
				axis background/.style={fill=white},%
				legend style={at={(0.50,0.85)},anchor=north,legend cell 
					align=left,draw=none,legend columns=-1,align=left,draw=white!15!black}
				]
				\addplot [color=blue, dotted, line width=0.75pt] 
				file{matlabPlots/Power_Tower_Inspection/acceleration_UAV42_x.txt};%
				\addplot [color=red, dashed, line width=0.75pt] 
				file{matlabPlots/Power_Tower_Inspection/acceleration_UAV42_y.txt};%
				\addplot [color=green, solid, line width=0.75pt] 
				file{matlabPlots/Power_Tower_Inspection/acceleration_UAV42_z.txt};%
				\draw [thick, dashed] (0, 3) -- (450, 3);
				\draw [thick, dashed] (0, -3) -- (450, -3);
				\draw [thick, dashed] (160, -4) -- (160, 4);
				\draw [thick, dashed] (320, -4) -- (320, 4);
				\fill[blue!50,nearly transparent] (0,3) -- (20,3) -- (20,-3) -- (0,-3) -- cycle;
				\fill[red!50,nearly transparent] (20,3) -- (40,3) -- (40,-3) -- (20,-3) -- 
				cycle;
				\fill[green!50,nearly transparent] (40,3) -- (60,3) -- (60,-3) -- (40,-3) -- 
				cycle;
				\legend{$a^{(1)}$, $a^{(2)}$, $a^{(3)}$};%
				\end{axis}
				\end{tikzpicture}
			}
		\end{subfigure}
		\hspace{0.25cm}
		\begin{subfigure}{0.45\columnwidth}
			\centering
			\scalebox{0.52}{
				\begin{tikzpicture}
				\begin{axis}[%
				width=2.8119in,%
				height=1.8183in,%
				at={(0.758in,0.481in)},%
				scale only axis,%
				xmin=0,%
				xmax=60,%
				ymax=4,%
				ymin=-4,%
				xmajorgrids,%
				ymajorgrids,%
				extra y ticks={-3,3},
				extra y tick labels={$-\mathbf{a}^{(j)}_\mathrm{max}$, 
				$\mathbf{a}^{(j)}_\mathrm{max}$},
				extra y tick style={ticklabel style={xshift=0pt}}, % if they crash with the 
				%default ticks
				%extra x ticks={20,40},
				%extra x tick labels={$\varphi_\mathrm{tr1} \text{,} \, \varphi_\mathrm{tr2}$, 
				%	$\varphi_\mathrm{tr3} \text{,} \, \varphi_\mathrm{tr4}$},
				%extra x tick style={ticklabel style={yshift=-10pt}}, % if they crash with the 
				%default ticks
				ylabel style={yshift=-0.775cm}, %shifting the y line text
				%xlabel={Time [\si{\second}]},%
				ylabel={[\si{\meter\per\square\second}]},%
				axis background/.style={fill=white},%
				legend style={at={(0.50,0.85)},anchor=north,legend cell 
					align=left,draw=none,legend columns=-1,align=left,draw=white!15!black}
				]
				\addplot [color=blue, dotted, line width=0.75pt] 
				file{matlabPlots/Power_Tower_Inspection/acceleration_UAV53_x.txt};%
				\addplot [color=red, dashed, line width=0.75pt] 
				file{matlabPlots/Power_Tower_Inspection/acceleration_UAV53_y.txt};%
				\addplot [color=green, solid, line width=0.75pt] 
				file{matlabPlots/Power_Tower_Inspection/acceleration_UAV53_z.txt};%
				\draw [thick, dashed] (0, 3) -- (450, 3);
				\draw [thick, dashed] (0, -3) -- (450, -3);
				\draw [thick, dashed] (160, -4) -- (160, 4);
				\draw [thick, dashed] (320, -4) -- (320, 4);
				\fill[blue!50,nearly transparent] (0,3) -- (20,3) -- (20,-3) -- (0,-3) -- cycle;
				\fill[red!50,nearly transparent] (20,3) -- (40,3) -- (40,-3) -- (20,-3) -- 
				cycle;
				\fill[green!50,nearly transparent] (40,3) -- (60,3) -- (60,-3) -- (40,-3) -- 
				cycle;
				\legend{$a^{(1)}$, $a^{(2)}$, $a^{(3)}$};%
				\end{axis}
				\end{tikzpicture}
			}
		\end{subfigure}
		\\
		\vspace{0.05cm}
		\hspace{-0.75cm}
		\begin{subfigure}{0.45\columnwidth}
			\centering
			\scalebox{0.52}{
				\begin{tikzpicture}
				\begin{axis}[%
				width=2.8119in,%
				height=1.8183in,%
				at={(0.758in,0.481in)},%
				scale only axis,%
				xmin=0,%
				xmax=60,%
				ymax=12,%
				ymin=0,%
				xmajorgrids,%
				ymajorgrids,%
				extra y ticks={3},
				extra y tick labels={$\delta_\mathrm{min}$},
				extra y tick style={ticklabel style={xshift=0pt}}, % if they crash with the 
				%default ticks
				extra x ticks={20,40},
				extra x tick labels={$\varphi_\mathrm{tr1} \text{,} \, \varphi_\mathrm{tr2}$, 
					$\varphi_\mathrm{tr3} \text{,} \, \varphi_\mathrm{tr4}$},
				extra x tick style={ticklabel style={yshift=-10pt}}, % if they crash with the 
				%default ticks
				ylabel style={yshift=-0.415cm}, %shifting the y line text
				xlabel={Time [\si{\second}]},%
				ylabel={[\si{\meter}]},%
				axis background/.style={fill=white},%
				legend style={at={(0.50,0.45)},anchor=north,legend cell 
					align=left,draw=none,legend columns=-1,align=left,draw=white!15!black}
				]
				\addplot [color=blue, solid, line width=0.75pt] 
				file{matlabPlots/Power_Tower_Inspection/mutual_distance.txt};%
				\draw [thick, dashed] (0, 3) -- (450, 3);
				\fill[blue!50,nearly transparent] (0,3) -- (20,3) -- (20,12) -- (0,12) -- cycle;
				\fill[red!50,nearly transparent] (20,3) -- (40,3) -- (40,12) -- (20,12) -- 
				cycle;
				\fill[green!50,nearly transparent] (40,3) -- (60,3) -- (60,12) -- (40,12) -- 
				cycle;
				\legend{$\lVert \prescript{1}{}{\mathbf{p}} - \prescript{2}{}{\mathbf{p}} \rVert$};%
				\end{axis}
				\end{tikzpicture}
			}
		\end{subfigure}
		\hspace{0.2cm}
		\begin{subfigure}{0.45\columnwidth}
			\centering
			\vspace{-0.85mm}
			\scalebox{0.52}{
				\begin{tikzpicture}
				\begin{axis}[%
				width=2.8119in,%
				height=1.8183in,%
				at={(0.758in,0.481in)},%
				scale only axis,%
				xmin=0,%
				xmax=60,%
				ymax=1,%
				ymin=0,%
				xmajorgrids,%
				ymajorgrids,%
				extra y tick style={ticklabel style={xshift=0pt}}, % if they crash with the 
				%default ticks
				extra x ticks={20,40},
				extra x tick labels={$\varphi_\mathrm{tr1} \text{,} \, \varphi_\mathrm{tr2}$, 
					$\varphi_\mathrm{tr3} \text{,} \, \varphi_\mathrm{tr4}$},
				extra x tick style={ticklabel style={yshift=-10pt}}, % if they crash with the 
				%default ticks
				ylabel style={yshift=-0.120cm}, %shifting the y line text
				xlabel={Time [\si{\second}]},%
				ylabel={Robust Semantic},%
				axis background/.style={fill=white},%
				legend style={at={(0.35,0.45)},anchor=north,legend cell 
					align=left,draw=none,legend columns=-1,align=left,draw=white!15!black}
				]
				\addplot [color=blue, dotted, line width=0.75pt] 
				file{matlabPlots/Power_Tower_Inspection/rho_goal_UAV53.txt};%
				\addplot [color=blue, dashed, line width=0.75pt] 
				file{matlabPlots/Power_Tower_Inspection/rho_goal_UAV42.txt};%
				\fill[blue!50,nearly transparent] (0,0) -- (20,0) -- (20,1) -- (0,1) -- cycle;
				\fill[red!50,nearly transparent] (20,0) -- (40,0) -- (40,1) -- (20,1) -- 
				cycle;
				\fill[green!50,nearly transparent] (40,0) -- (60,0) -- (60,1) -- (40,1) -- 
				cycle;
				\legend{$\rho_\mathrm{drone1}$, $\rho_\mathrm{drone2}$};%
				\end{axis}
				\end{tikzpicture}
			}
		\end{subfigure}
	\end{center}
	\vspace{-3.5mm}
	\caption{Position, linear velocity and acceleration, and mission requirements considering the 
	``basic'' motion planner. From left to 	right: ``drone1'' and ``drone2'' data. The STL 
	specifications ($\varphi_\mathrm{tr1}$, $\varphi_\mathrm{tr2}$, $\varphi_\mathrm{tr3}$, and 
	$\varphi_\mathrm{tr4}$) are also reported with different color regions.}
	\label{fig:graphConstraintsPowerTower}
  \vspace{-1em}
\end{figure}
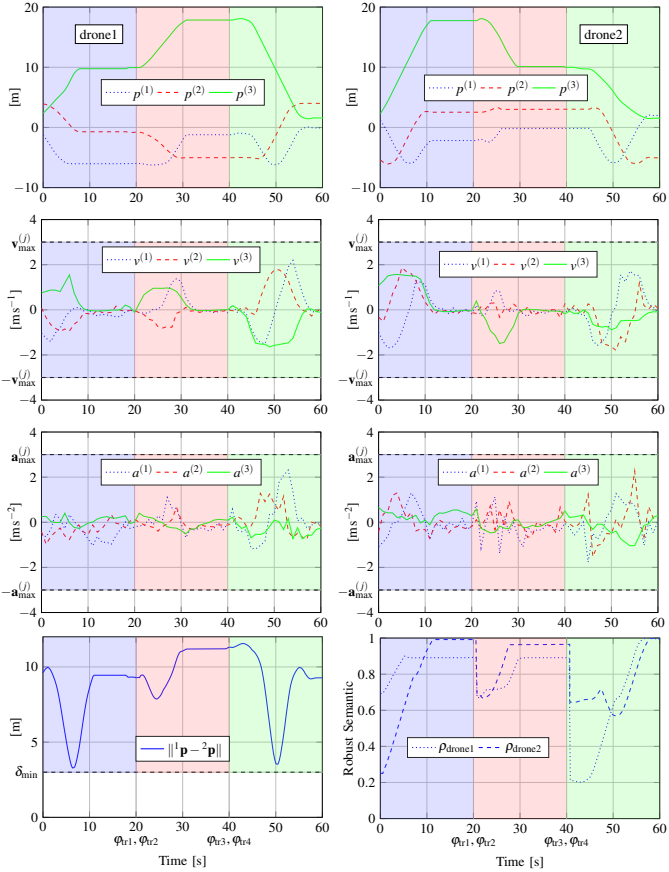

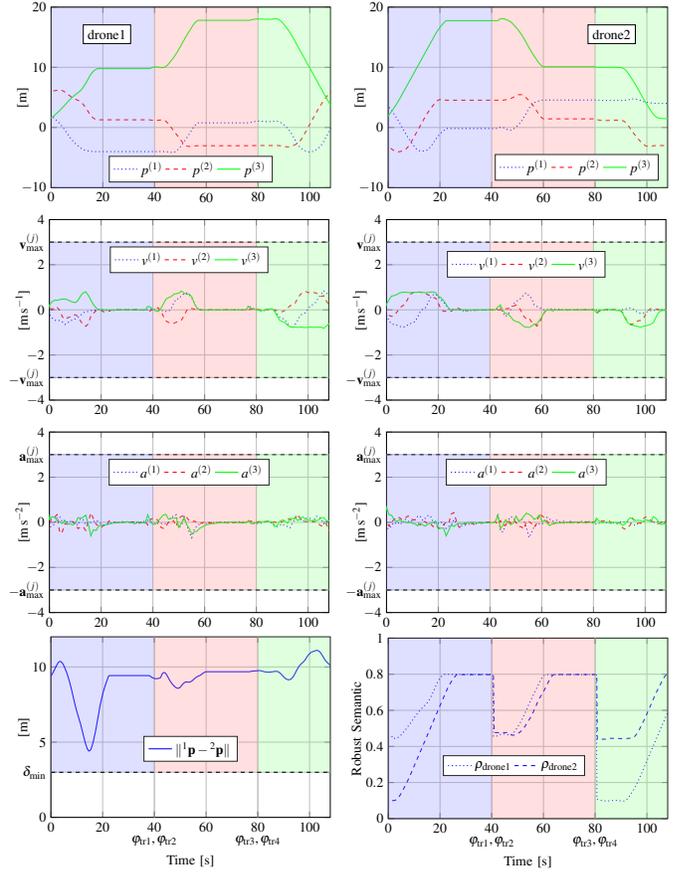
\begin{figure}
	\vspace{-1.75em}
	\begin{center}
		\hspace{-0.725cm}
		\begin{subfigure}{0.45\columnwidth}
			\scalebox{0.52}{
				\begin{tikzpicture}
				\begin{axis}[%
				width=2.8119in,%
				height=1.8183in,%
				at={(0.758in,0.481in)},%
				scale only axis,%
				xmin=0,%
				xmax=108,%
				ymax=20,%
				ymin=-10,%
				xmajorgrids,%
				ymajorgrids,%
				ylabel style={yshift=-0.415cm}, %shifting the y line text
				%xlabel={Time [\si{\second}]},%
				ylabel={[\si{\meter}]},%
				axis background/.style={fill=white},%
				legend style={at={(0.50,0.175)},anchor=north,legend cell 
					align=left,draw=none,legend columns=-1,align=left,draw=white!15!black}
				]
				\addplot [color=blue, dotted, line width=0.75pt] 
				file{matlabPlots/Power_Tower_Energy/position_UAV1_x.txt};%
				\addplot [color=red, dashed, line width=0.75pt] 
				file{matlabPlots/Power_Tower_Energy/position_UAV1_y.txt};%
				\addplot [color=green, solid, line width=0.75pt] 
				file{matlabPlots/Power_Tower_Energy/position_UAV1_z.txt};%
				\draw [thick, dashed] (160, -10) -- (160, 20);
				\draw [thick, dashed] (320, -10) -- (320, 20);
				\fill[blue!50,nearly transparent] (0,20) -- (40,20) -- (40,-10) -- (0,-10) -- 
				cycle;
				\fill[red!50,nearly transparent] (40,20) -- (80,20) -- (80,-10) -- (40,-10) -- 
				cycle;
				\fill[green!50,nearly transparent] (80,20) -- (120,20) -- (120,-10) -- (80,-10) 
				-- cycle;
				\legend{$p^{(1)}$, $p^{(2)}$, $p^{(3)}$};%
				\node [draw,fill=white] at (rel axis cs: 0.2,0.85) {\shortstack[l]{drone1}};
				\end{axis}
				\end{tikzpicture}
			}
		\end{subfigure}
		\hspace{0.25cm}
		\begin{subfigure}{0.45\columnwidth}
			\scalebox{0.52}{
				\begin{tikzpicture}
				\begin{axis}[%
				width=2.8119in,%
				height=1.8183in,%
				at={(0.758in,0.481in)},%
				scale only axis,%
				xmin=0,%
				xmax=108,%
				ymax=20,%
				ymin=-10,%
				xmajorgrids,%
				ymajorgrids,%
				ylabel style={yshift=-0.415cm}, %shifting the y line text
				%xlabel={Time [\si{\second}]},%
				ylabel={[\si{\meter}]},%
				axis background/.style={fill=white},%
				legend style={at={(0.675,0.185)},anchor=north,legend cell 
					align=left,draw=none,legend columns=-1,align=left,draw=white!15!black}
				]
				\addplot [color=blue, dotted, line width=0.75pt] 				
				file{matlabPlots/Power_Tower_Energy/position_UAV2_x.txt};%
				\addplot [color=red, dashed, line width=0.75pt] 
				file{matlabPlots/Power_Tower_Energy/position_UAV2_y.txt};%
				\addplot [color=green, solid, line width=0.75pt] 
				file{matlabPlots/Power_Tower_Energy/position_UAV2_z.txt};%
				\draw [thick, dashed] (160, -10) -- (160, 20);
				\draw [thick, dashed] (320, -10) -- (320, 20);
				\fill[blue!50,nearly transparent] (0,20) -- (40,20) -- (40,-10) -- (0,-10) -- 
				cycle;
				\fill[red!50,nearly transparent] (40,20) -- (80,20) -- (80,-10) -- (40,-10) -- 
				cycle;
				\fill[green!50,nearly transparent] (80,20) -- (120,20) -- (120,-10) -- (80,-10) 
				-- cycle;
				\legend{$p^{(1)}$, $p^{(2)}$, $p^{(3)}$};%
				\node [draw,fill=white] at (rel axis cs: 0.8,0.85) {\shortstack[l]{drone2}};
				\end{axis}
				\end{tikzpicture}
			}
		\end{subfigure}
		\\
		\vspace{0.05cm}
		\hspace{-0.95cm}
		\begin{subfigure}{0.45\columnwidth}
			\centering
			\scalebox{0.52}{
				\begin{tikzpicture}
				\begin{axis}[%
				width=2.8119in,%
				height=1.8183in,%
				at={(0.758in,0.481in)},%
				scale only axis,%
				xmin=0,%
				xmax=108,%
				ymax=4,%
				ymin=-4,%
				xmajorgrids,%
				ymajorgrids,%
				extra y ticks={-3,3},
				extra y tick labels={$-\mathbf{v}^{(j)}_\mathrm{max}$, 
					$\mathbf{v}^{(j)}_\mathrm{max}$},
				extra y tick style={ticklabel style={xshift=0pt}}, % if they crash with the 
				%default ticks
				ylabel style={yshift=-0.775cm}, %shifting the y line text
				%xlabel={Time [\si{\second}]},%
				ylabel={[\si{\meter\per\second}]},%
				axis background/.style={fill=white},%
				legend style={at={(0.50,0.85)},anchor=north,legend cell 
					align=left,draw=none,legend columns=-1,align=left,draw=white!15!black}
				]
				\addplot [color=blue, dotted, line width=0.75pt] 
				file{matlabPlots/Power_Tower_Energy/velocity_UAV1_x.txt};%
				\addplot [color=red, dashed, line width=0.75pt] 
				file{matlabPlots/Power_Tower_Energy/velocity_UAV1_y.txt};%
				\addplot [color=green, solid, line width=0.75pt] 
				file{matlabPlots/Power_Tower_Energy/velocity_UAV1_z.txt};%
				\draw [thick, dashed] (0, 3) -- (450, 3);
				\draw [thick, dashed] (0, -3) -- (450, -3);
				\draw [thick, dashed] (160, -4) -- (160, 4);
				\draw [thick, dashed] (320, -4) -- (320, 4);
				\fill[blue!50,nearly transparent] (0,3) -- (40,3) -- (40,-3) -- (0,-3) -- 
				cycle;
				\fill[red!50,nearly transparent] (40,3) -- (80,3) -- (80,-3) -- (40,-3) -- 
				cycle;
				\fill[green!50,nearly transparent] (80,3) -- (120,3) -- (120,-3) -- (80,-3) 
				-- cycle;
				\legend{$v^{(1)}$, $v^{(2)}$, $v^{(3)}$};%
				\end{axis}
				\end{tikzpicture}
			}
		\end{subfigure}
		\hspace{0.25cm}
		\begin{subfigure}{0.45\columnwidth}
			\centering
			\scalebox{0.52}{
				\begin{tikzpicture}
				\begin{axis}[%
				width=2.8119in,%
				height=1.8183in,%
				at={(0.758in,0.481in)},%
				scale only axis,%
				xmin=0,%
				xmax=108,%
				ymax=4,%
				ymin=-4,%
				xmajorgrids,%
				ymajorgrids,%
				extra y ticks={-3,3},
				extra y tick labels={$-\mathbf{v}^{(j)}_\mathrm{max}$, 
					$\mathbf{v}^{(j)}_\mathrm{max}$},
				extra y tick style={ticklabel style={xshift=0pt}}, % if they crash with the 
				%default ticks
				ylabel style={yshift=-0.775cm}, %shifting the y line text
				%xlabel={Time [\si{\second}]},%
				ylabel={[\si{\meter\per\second}]},%
				axis background/.style={fill=white},%
				legend style={at={(0.50,0.825)},anchor=north,legend cell 
					align=left,draw=none,legend columns=-1,align=left,draw=white!15!black}
				]
				\addplot [color=blue, dotted, line width=0.75pt] 
				file{matlabPlots/Power_Tower_Energy/velocity_UAV2_x.txt};%
				\addplot [color=red, dashed, line width=0.75pt] 
				file{matlabPlots/Power_Tower_Energy/velocity_UAV2_y.txt};%
				\addplot [color=green, solid, line width=0.75pt] 
				file{matlabPlots/Power_Tower_Energy/velocity_UAV2_z.txt};%
				\draw [thick, dashed] (0, 3) -- (450, 3);
				\draw [thick, dashed] (0, -3) -- (450, -3);
				\draw [thick, dashed] (160, -4) -- (160, 4);
				\draw [thick, dashed] (320, -4) -- (320, 4);
				\fill[blue!50,nearly transparent] (0,3) -- (40,3) -- (40,-3) -- (0,-3) -- 
				cycle;
				\fill[red!50,nearly transparent] (40,3) -- (80,3) -- (80,-3) -- (40,-3) -- 
				cycle;
				\fill[green!50,nearly transparent] (80,3) -- (120,3) -- (120,-3) -- (80,-3) 
				-- cycle;
				\legend{$v^{(1)}$, $v^{(2)}$, $v^{(3)}$};%
				\end{axis}
				\end{tikzpicture}
			}
		\end{subfigure}
		\\
		\vspace{0.05cm}
		\hspace{-0.95cm}
		\begin{subfigure}{0.45\columnwidth}
			\centering
			\scalebox{0.52}{
				\begin{tikzpicture}
				\begin{axis}[%
				width=2.8119in,%
				height=1.8183in,%
				at={(0.758in,0.481in)},%
				scale only axis,%
				xmin=0,%
				xmax=108,%
				ymax=4,%
				ymin=-4,%
				xmajorgrids,%
				ymajorgrids,%
				extra y ticks={-3,3},
				extra y tick labels={$-\mathbf{a}^{(j)}_\mathrm{max}$, 
					$\mathbf{a}^{(j)}_\mathrm{max}$},
				extra y tick style={ticklabel style={xshift=0pt}}, % if they crash with the 
				%default ticks
				%extra x ticks={20,40},
				%extra x tick labels={$\varphi_\mathrm{tr1} \text{,} \, \varphi_\mathrm{tr2}$, 
				%	$\varphi_\mathrm{tr3} \text{,} \, \varphi_\mathrm{tr4}$},
				%extra x tick style={ticklabel style={yshift=-10pt}}, % if they crash with the 
				%default ticks
				ylabel style={yshift=-0.775cm}, %shifting the y line text
				%xlabel={Time [\si{\second}]},%
				ylabel={[\si{\meter\per\square\second}]},%
				axis background/.style={fill=white},%
				legend style={at={(0.50,0.85)},anchor=north,legend cell 
					align=left,draw=none,legend columns=-1,align=left,draw=white!15!black}
				]
				\addplot [color=blue, dotted, line width=0.75pt] 
				file{matlabPlots/Power_Tower_Energy/acceleration_UAV1_x.txt};%
				\addplot [color=red, dashed, line width=0.75pt] 
				file{matlabPlots/Power_Tower_Energy/acceleration_UAV1_y.txt};%
				\addplot [color=green, solid, line width=0.75pt] 
				file{matlabPlots/Power_Tower_Energy/acceleration_UAV1_z.txt};%
				\draw [thick, dashed] (0, 3) -- (450, 3);
				\draw [thick, dashed] (0, -3) -- (450, -3);
				\draw [thick, dashed] (160, -4) -- (160, 4);
				\draw [thick, dashed] (320, -4) -- (320, 4);
				\fill[blue!50,nearly transparent] (0,3) -- (40,3) -- (40,-3) -- (0,-3) -- 
				cycle;
				\fill[red!50,nearly transparent] (40,3) -- (80,3) -- (80,-3) -- (40,-3) -- 
				cycle;
				\fill[green!50,nearly transparent] (80,3) -- (120,3) -- (120,-3) -- (80,-3) 
				-- cycle;
				\legend{$a^{(1)}$, $a^{(2)}$, $a^{(3)}$};%
				\end{axis}
				\end{tikzpicture}
			}
		\end{subfigure}
		\hspace{0.25cm}
		\begin{subfigure}{0.45\columnwidth}
			\centering
			\scalebox{0.52}{
				\begin{tikzpicture}
				\begin{axis}[%
				width=2.8119in,%
				height=1.8183in,%
				at={(0.758in,0.481in)},%
				scale only axis,%
				xmin=0,%
				xmax=108,%
				ymax=4,%
				ymin=-4,%
				xmajorgrids,%
				ymajorgrids,%
				extra y ticks={-3,3},
				extra y tick labels={$-\mathbf{a}^{(j)}_\mathrm{max}$, 
					$\mathbf{a}^{(j)}_\mathrm{max}$},
				extra y tick style={ticklabel style={xshift=0pt}}, % if they crash with the 
				%default ticks
				%extra x ticks={20,40},
				%extra x tick labels={$\varphi_\mathrm{tr1} \text{,} \, \varphi_\mathrm{tr2}$, 
				%	$\varphi_\mathrm{tr3} \text{,} \, \varphi_\mathrm{tr4}$},
				%extra x tick style={ticklabel style={yshift=-10pt}}, % if they crash with the 
				%default ticks
				ylabel style={yshift=-0.775cm}, %shifting the y line text
				%xlabel={Time [\si{\second}]},%
				ylabel={[\si{\meter\per\square\second}]},%
				axis background/.style={fill=white},%
				legend style={at={(0.50,0.85)},anchor=north,legend cell 
					align=left,draw=none,legend columns=-1,align=left,draw=white!15!black}
				]
				\addplot [color=blue, dotted, line width=0.75pt] 
				file{matlabPlots/Power_Tower_Energy/acceleration_UAV2_x.txt};%
				\addplot [color=red, dashed, line width=0.75pt] 
				file{matlabPlots/Power_Tower_Energy/acceleration_UAV2_y.txt};%
				\addplot [color=green, solid, line width=0.75pt] 
				file{matlabPlots/Power_Tower_Energy/acceleration_UAV2_z.txt};%
				\draw [thick, dashed] (0, 3) -- (450, 3);
				\draw [thick, dashed] (0, -3) -- (450, -3);
				\draw [thick, dashed] (160, -4) -- (160, 4);
				\draw [thick, dashed] (320, -4) -- (320, 4);
				\fill[blue!50,nearly transparent] (0,3) -- (40,3) -- (40,-3) -- (0,-3) -- 
				cycle;
				\fill[red!50,nearly transparent] (40,3) -- (80,3) -- (80,-3) -- (40,-3) -- 
				cycle;
				\fill[green!50,nearly transparent] (80,3) -- (120,3) -- (120,-3) -- (80,-3) 
				-- cycle;
				\legend{$a^{(1)}$, $a^{(2)}$, $a^{(3)}$};%
				\end{axis}
				\end{tikzpicture}
			}
		\end{subfigure}
		\\
		\vspace{0.05cm}
		\hspace{-0.75cm}
		\begin{subfigure}{0.45\columnwidth}
			\centering
			\scalebox{0.52}{
				\begin{tikzpicture}
				\begin{axis}[%
				width=2.8119in,%
				height=1.8183in,%
				at={(0.758in,0.481in)},%
				scale only axis,%
				xmin=0,%
				xmax=108,%
				ymax=12,%
				ymin=0,%
				xmajorgrids,%
				ymajorgrids,%
				extra y ticks={3},
				extra y tick labels={$\delta_\mathrm{min}$},
				extra y tick style={ticklabel style={xshift=0pt}}, % if they crash with the 
				%default ticks
				extra x ticks={40,80},
				extra x tick labels={$\varphi_\mathrm{tr1} \text{,} \, \varphi_\mathrm{tr2}$, 
					$\varphi_\mathrm{tr3} \text{,} \, \varphi_\mathrm{tr4}$},
				extra x tick style={ticklabel style={yshift=-10pt}}, % if they crash with the 
				%default ticks
				ylabel style={yshift=-0.415cm}, %shifting the y line text
				xlabel={Time [\si{\second}]},%
				ylabel={[\si{\meter}]},%
				axis background/.style={fill=white},%
				legend style={at={(0.50,0.45)},anchor=north,legend cell 
					align=left,draw=none,legend columns=-1,align=left,draw=white!15!black}
				]
				\addplot [color=blue, solid, line width=0.75pt] 
				file{matlabPlots/Power_Tower_Energy/mutual_distance.txt};%
				\draw [thick, dashed] (0, 3) -- (108, 3);
				\fill[blue!50,nearly transparent] (0,3) -- (40,3) -- (40,15) -- (0,15) -- 
				cycle;
				\fill[red!50,nearly transparent] (40,3) -- (80,3) -- (80,15) -- (40,15) -- 
				cycle;
				\fill[green!50,nearly transparent] (80,3) -- (120,3) -- (120,15) -- (80,15) 
				-- cycle;
				\legend{$\lVert \prescript{1}{}{\mathbf{p}} - \prescript{2}{}{\mathbf{p}} \rVert$};%
				\end{axis}
				\end{tikzpicture}
			}
		\end{subfigure}
		\hspace{0.2cm}
		\begin{subfigure}{0.45\columnwidth}
			\centering
			\vspace{-0.85mm}
			\scalebox{0.52}{
				\begin{tikzpicture}
				\begin{axis}[%
				width=2.8119in,%
				height=1.8183in,%
				at={(0.758in,0.481in)},%
				scale only axis,%
				xmin=0,%
				xmax=108,%
				ymax=1,%
				ymin=0,%
				xmajorgrids,%
				ymajorgrids,%
				extra y tick style={ticklabel style={xshift=0pt}}, % if they crash with the 
				%default ticks
				extra x ticks={40,80},
				extra x tick labels={$\varphi_\mathrm{tr1} \text{,} \, \varphi_\mathrm{tr2}$, 
					$\varphi_\mathrm{tr3} \text{,} \, \varphi_\mathrm{tr4}$},
				extra x tick style={ticklabel style={yshift=-10pt}}, % if they crash with the 
				%default ticks
				ylabel style={yshift=-0.120cm}, %shifting the y line text
				xlabel={Time [\si{\second}]},%
				ylabel={Robust Semantic},%
				axis background/.style={fill=white},%
				legend style={at={(0.45,0.35)},anchor=north,legend cell 
					align=left,draw=none,legend columns=-1,align=left,draw=white!15!black}
				]
				\addplot [color=blue, dotted, line width=0.75pt] 
				file{matlabPlots/Power_Tower_Energy/rho_goal_UAV2.txt};%
				\addplot [color=blue, dashed, line width=0.75pt] 
				file{matlabPlots/Power_Tower_Energy/rho_goal_UAV1.txt};%
				\fill[blue!50,nearly transparent] (0,0) -- (40,0) -- (40,1) -- (0,1) -- 
				cycle;
				\fill[red!50,nearly transparent] (40,0) -- (80,0) -- (80,1) -- (40,1) -- 
				cycle;
				\fill[green!50,nearly transparent] (80,0) -- (120,0) -- (120,1) -- (80,1) 
				-- cycle;
				\legend{$\rho_\mathrm{drone1}$, $\rho_\mathrm{drone2}$};%
				\end{axis}
				\end{tikzpicture}
			}
		\end{subfigure}
	\end{center}
	\vspace{-3.5mm}
	\caption{Position, linear velocity and acceleration and, mission requirements when considering 
	the energy-aware motion planner performing the \textit{power tower inspection}.}
	\label{fig:energyAccelerationRobustness}
\end{figure}

As for the previous task, a pair of quad-rotors performs an inspection of a single power 
tower. The \textit{power tower inspection} scenario was considered to also evaluate the performance 
of the \textit{energy-aware} and \textit{event-triggering} replanner. The results of the numerical 
simulations carried out in MATLAB are reported in 
Figs.~\ref{fig:powerTowerInspectionEnergyMinimization} 
and~\ref{fig:powerTowerEventTriggeredReplanner}. As expected, the trajectories obtained considering 
the energy requirements (see Fig.~\ref{fig:powerTowerInspectionEnergyMinimization}) result closer 
to the obstacles than what happens when no energy requirement is enforced (see 
Fig.~\ref{fig:powerTowerInspection}). This fact is motivated by the introduction of the energy 
saving cost term, at the expense of the overall robustness (see 
Figs.~\ref{fig:graphConstraintsPowerTower} and~\ref{fig:energyAccelerationRobustness}).

To validate the performance of the event-triggered replanner, we simulated the presence of 
two major unexpected disturbances deviating at runtime the quad-rotors from their original planned 
paths. Once the replanner detects major deviations from the planned trajectory (i.e., $\lvert 
\tilde{\mathbf{p}}_l - \mathbf{p}_l \rvert > \eta$), a partial replanning is triggered online to 
bring back the quad-rotors to next ``topic'' waypoint, as illustrated in 
Fig.~\ref{fig:powerTowerEventTriggeredReplanner}. Then, the result is used as reference for 
the tracking controller. The optimization took less than \SI{1}{\second} for both disturbances. 

\begin{figure}
	\begin{center}
		\hspace{-0.825cm}
			\begin{subfigure}{0.45\columnwidth}
			\centering
			\scalebox{0.52}{
				\begin{tikzpicture}
				\begin{axis}[%
				width=2.8119in,%
				height=1.8183in,%
				at={(0.758in,0.481in)},%
				scale only axis,%
				xmin=0,%
				xmax=60,%
				ymax=8,%
				ymin=0,%
				xmajorgrids,%
				ymajorgrids,%
				extra y tick style={ticklabel style={xshift=0pt}}, % if they crash with the 
				%default ticks
				extra x ticks={20,40},
				extra x tick labels={$\varphi_\mathrm{tr1} \text{,} \, \varphi_\mathrm{tr2}$, 
					$\varphi_\mathrm{tr3} \text{,} \, \varphi_\mathrm{tr4}$},
				extra x tick style={ticklabel style={yshift=-10pt}}, % if they crash with the 
				%default ticks
				ylabel style={yshift=0cm}, %shifting the y line text
				xlabel={Time [\si{\second}]},%
				ylabel={[\si{\joule}]},%
				axis background/.style={fill=white},%
				legend style={at={(0.50,0.85)},anchor=north,legend cell 
					align=left,draw=none,legend columns=-1,align=left,draw=white!15!black}
				]
				\addplot [color=blue, dotted, line width=0.75pt] 
				file{matlabPlots/Power_Tower_Inspection/norm_UAV42_PT.txt};%
				\addplot [color=red, dashed, line width=0.75pt] 
				file{matlabPlots/Power_Tower_Inspection/norm_UAV53_PT.txt};%
				\legend{$\mathrm{drone1}$, $\mathrm{drone2}$};%
				\end{axis}
				\end{tikzpicture}
			}
		\end{subfigure}
		\hspace{0.35cm}
		\begin{subfigure}{0.45\columnwidth}
			\centering
			\scalebox{0.52}{
				\begin{tikzpicture}
				\begin{axis}[%
				width=2.8119in,%
				height=1.8183in,%
				at={(0.758in,0.481in)},%
				scale only axis,%
				xmin=0,%
				xmax=108,%
				ymax=1,%
				ymin=0,%
				xmajorgrids,%
				ymajorgrids,%
				extra y tick style={ticklabel style={xshift=0pt}}, % if they crash with the 
				%default ticks
				%extra y ticks={0.0754,0.0631},
				%extra y tick labels={$\mathbf{E}_\mathrm{drone1}$, 				
				%$\mathbf{E}_\mathrm{drone1}$},
				extra x ticks={40,80},
				extra x tick labels={$\varphi_\mathrm{tr1} \text{,} \, \varphi_\mathrm{tr2}$, 
					$\varphi_\mathrm{tr3} \text{,} \, \varphi_\mathrm{tr4}$},
				extra x tick style={ticklabel style={yshift=-10pt}}, % if they crash with the 
				%default ticks
				ylabel style={yshift=0cm}, %shifting the y line text
				xlabel={Time [\si{\second}]},%
				ylabel={[\si{\joule}]},%
				axis background/.style={fill=white},%
				legend style={at={(0.50,0.95)},anchor=north,legend cell 
					align=left,draw=none,legend columns=-1,align=left,draw=white!15!black}
				]
				\addplot [color=blue, dotted, line width=0.75pt] 
				file{matlabPlots/Power_Tower_Energy/norm_UAV1_EA.txt};%
				\addplot [color=red, dashed, line width=0.75pt] 
				file{matlabPlots/Power_Tower_Energy/norm_UAV2_EA.txt};%
				\legend{$\mathrm{drone1}$, $\mathrm{drone2}$};%
				\end{axis}
				\end{tikzpicture}
			}
		\end{subfigure}
	\end{center}
	\vspace{-2.5mm}
	\caption{Energy consumption profiles with two quad-rotors performing the \textit{power tower 
	inspection}. From left to right: ``drone1'' and ``drone2'' data when considering the 
	``basic''~\eqref{eq:optimizationProblemMotionPrimitives} and the 	
	energy-aware~\eqref{eq:energeryAwareMotionPlanner} motion planner, respectively.}
	\label{fig:comparisionEnergyScenarios}
\end{figure}
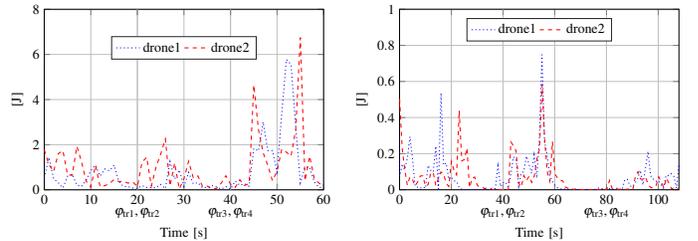

Gazebo simulations were performed to evaluate the decrease in energy when using the 	
trajectories obtained with the energy-aware motion planner, as shown in 
Fig.~\ref{fig:energyAccelerationRobustness}. As can be seen from the graph, the velocity and 
acceleration signals are still within the bounds and assume lower pick-values than 
Fig.~\ref{fig:graphConstraintsPowerTower}. However, smaller values of the robustness are obtained. 
In Fig.~\ref{fig:comparisionEnergyScenarios} the energy consumption profiles are reported 
for the two problem formulations.

%\begin{table}
%	\centering
%	\begin{adjustbox}{max width=\columnwidth}
%		\begin{tabular}{|l|c|c|c|c|}
%			\hline
%			& \textbf{Sym.} & \textbf{Value-EA} &  \textbf{Value-PT} & \textbf{Unit} 
%			\\
%			\hline
%			Drone1 & $\prescript{1}{}{E}$ &  &  & \si{\joule} \\
%			Drone2 & $\prescript{2}{}{E}$ &  &  & \si{\joule} \\
%			Period & $T$ & 110 & 60 & \si{\second} \\
%			\hline
%		\end{tabular}
%	\end{adjustbox}
%	\caption{The sum of the energy profiles considering the energy-aware and ``basic'' motion 
%	planner.}
%	\label{tab:integralEnergy}
%\end{table}

%%% END SECTION ============================================================

%%% START SECTION ==========================================================

\subsection{Comparison with kinodynamic RRT\textsuperscript{$\star$}}
\label{sec:kinodynamicRRT}

\begin{figure}
	\begin{center}
		%\hspace*{-12mm}
		% left - bottom - right - top
		\adjincludegraphics[trim={{.075\width} {.0\height} {0.0\width}				
		{.0\height}},clip,scale=0.205]{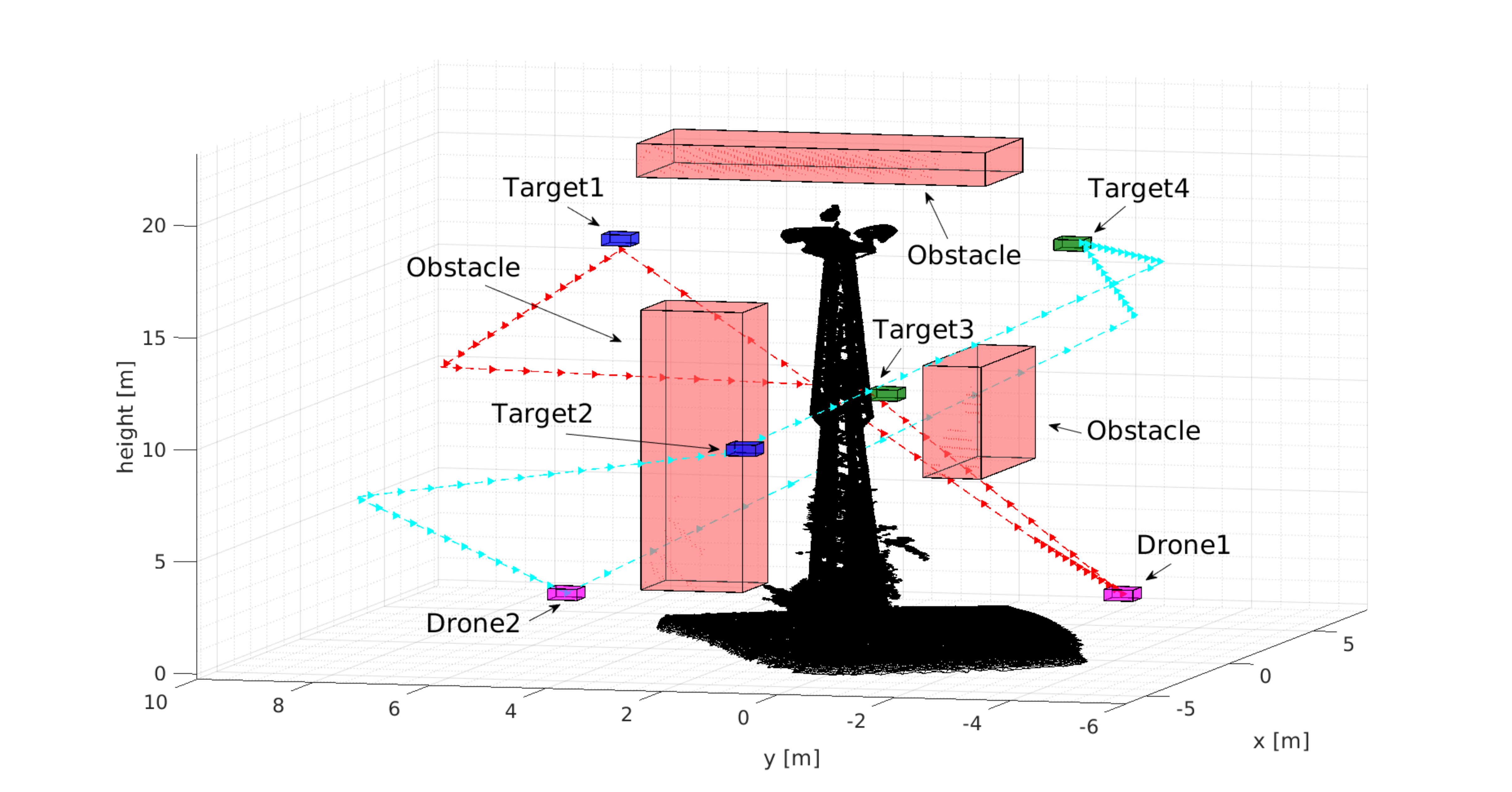}
		\vspace*{-9mm}
	\end{center}
	\caption{\textit{Power tower inspection} scenario when considering the kinodynamic 
	RRT\textsuperscript{$\star$}.}
	\label{fig:kinodynamicRRT}
	\vspace{-1em}
\end{figure}

As described in Sec.~\ref{sec:relatedWorks}, various state-of-the-art solutions investigate the 
path planning problem in quad-rotors inspection scenarios. However, not all of them are suitable 
for the inspection of power line infrastructure. This section aims to compare the set up 
optimization problem~\eqref{eq:optimizationProblemMotionPrimitives} with the kinodynamic 
RRT\textsuperscript{$\star$} proposed in~\cite{Webb2013ICRA}. Analogously to what done in our 
paper, the incremental sampling approach in~\cite{Webb2013ICRA} finds quad-rotor trajectories so 
as to remain in the workspace, avoid obstacles and incorporate bounds on the control inputs. 
Analogously to~\eqref{eq:splines}, the optimal trajectories are derived in terms of a solution of a 
$2n^2$-degree polynomial. In this case the dynamics of the quad-rotor is linearized around the 
hovering state constraining the yaw (and its derivatives) to zero~\cite[eq.~(31)]{Webb2013ICRA}. 
The obtained trajectories are reported in Fig.~\ref{fig:kinodynamicRRT}.

\begin{figure}
	\begin{center}
		\hspace{-0.725cm}
		\begin{subfigure}{0.45\columnwidth}
			\scalebox{0.52}{
				\begin{tikzpicture}
				\begin{axis}[%
				width=2.8119in,%
				height=1.8183in,%
				at={(0.758in,0.481in)},%
				scale only axis,%
				xmin=0,%
				xmax=80,%
				ymax=20,%
				ymin=-10,%
				xmajorgrids,%
				ymajorgrids,%
				extra x ticks={28,40},
				extra x tick labels={$\varphi_\mathrm{tr1} \text{,} \, \varphi_\mathrm{tr2}$, 
					~~~~$\varphi_\mathrm{tr3} \text{,} \, \varphi_\mathrm{tr4}$},
				extra x tick style={ticklabel style={yshift=-10pt}}, % if they crash with the 
				%default ticks
				ylabel style={yshift=-0.415cm}, %shifting the y line text
				xlabel={Time [\si{\second}]},%
				ylabel={[\si{\meter}]},%
				axis background/.style={fill=white},%
				legend style={at={(0.50,0.15)},anchor=north,legend cell 
					align=left,draw=none,legend columns=-1,align=left,draw=white!15!black}
				]
				\addplot [color=blue, dashed, line width=0.75pt] 
				file{matlabPlots/Kinodynamic_RRT/position_UAV1_x.txt};%
				\addplot [color=red, dashed, line width=0.75pt] 
				file{matlabPlots/Kinodynamic_RRT/position_UAV1_y.txt};%
				\addplot [color=green, dashed, line width=0.75pt] 
				file{matlabPlots/Kinodynamic_RRT/position_UAV1_z.txt};%
				\addplot [color=blue, solid, line width=0.75pt] 
				file{matlabPlots/Kinodynamic_RRT/position_UAV1_x_desired.txt};%
				\addplot [color=red, solid, line width=0.75pt] 
				file{matlabPlots/Kinodynamic_RRT/position_UAV1_y_desired.txt};%
				\addplot [color=green, solid, line width=0.75pt] 
				file{matlabPlots/Kinodynamic_RRT/position_UAV1_z_desired.txt};%
				\draw [thick, dashed] (160, -10) -- (160, 20);
				\draw [thick, dashed] (320, -10) -- (320, 20);
				\fill[blue!50,nearly transparent] (0,20) -- (28,20) -- (28,-10) -- (0,-10) -- 
				cycle;
				\fill[red!50,nearly transparent] (28,20) -- (40,20) -- (40,-10) -- (28,-10) -- 
				cycle;
				\fill[green!50,nearly transparent] (40,20) -- (78,20) -- (78,-10) -- (40,-10) 
				-- cycle;
				\legend{$p^{(1)}$, $p^{(2)}$, $p^{(3)}$};%
				\node [draw,fill=white] at (rel axis cs: 0.2,0.85) {\shortstack[l]{drone1}};
				\end{axis}
				\end{tikzpicture}
			}
		\end{subfigure}
		\hspace{0.25cm}
		\begin{subfigure}{0.45\columnwidth}
			\scalebox{0.52}{
				\begin{tikzpicture}
				\begin{axis}[%
				width=2.8119in,%
				height=1.8183in,%
				at={(0.758in,0.481in)},%
				scale only axis,%
				xmin=0,%
				xmax=80,%
				ymax=20,%
				ymin=-10,%
				xmajorgrids,%
				ymajorgrids,%
				extra x ticks={29,52},
				extra x tick labels={$\varphi_\mathrm{tr1} \text{,} \, \varphi_\mathrm{tr2}$, 
					$\varphi_\mathrm{tr3} \text{,} \, \varphi_\mathrm{tr4}$},
				extra x tick style={ticklabel style={yshift=-10pt}}, % if they crash with the 
				%default ticks
				ylabel style={yshift=-0.415cm}, %shifting the y line text
				xlabel={Time [\si{\second}]},%
				ylabel={[\si{\meter}]},%
				axis background/.style={fill=white},%
				legend style={at={(0.425,0.20)},anchor=north,legend cell 
					align=left,draw=none,legend columns=-1,align=left,draw=white!15!black}
				]
				\addplot [color=blue, dashed, line width=0.75pt] 
				file{matlabPlots/Kinodynamic_RRT/position_UAV2_x.txt};%
				\addplot [color=red, dashed, line width=0.75pt] 
				file{matlabPlots/Kinodynamic_RRT/position_UAV2_y.txt};%
				\addplot [color=green, dashed, line width=0.75pt] 
				file{matlabPlots/Kinodynamic_RRT/position_UAV2_z.txt};%
				\addplot [color=blue, solid, line width=0.75pt] 
				file{matlabPlots/Kinodynamic_RRT/position_UAV2_x_desired.txt};%
				\addplot [color=red, solid, line width=0.75pt] 
				file{matlabPlots/Kinodynamic_RRT/position_UAV2_y_desired.txt};%
				\addplot [color=green, solid, line width=0.75pt] 
				file{matlabPlots/Kinodynamic_RRT/position_UAV2_z_desired.txt};%
				\draw [thick, dashed] (160, -10) -- (160, 20);
				\draw [thick, dashed] (320, -10) -- (320, 20);
				\fill[blue!50,nearly transparent] (0,20) -- (29,20) -- (29,-10) -- (0,-10) -- 
				cycle;
				\fill[red!50,nearly transparent] (29,20) -- (52,20) -- (52,-10) -- (29,-10) -- 
				cycle;
				\fill[green!50,nearly transparent] (52,20) -- (68,20) -- (68,-10) -- (52,-10) 
				-- cycle;
				\legend{$p^{(1)}$, $p^{(2)}$, $p^{(3)}$};%
				\node [draw,fill=white] at (rel axis cs: 0.8,0.85) {\shortstack[l]{drone2}};
				\end{axis}
				\end{tikzpicture}
			}
		\end{subfigure}
	\end{center}
	\vspace{-2.5mm}
	\caption{Drone positions when considering the kinodynamic RRT\textsuperscript{$\star$}. Solid 
	lines represent the computed path, while dashed lines are the trajectories performed by the 
	quad-rotors. Color regions help understand when a part of the mission is accomplished in terms 
	of the corresponding STL case.}
	\label{fig:comparisonRRTDroneTrajGazebo}
\end{figure}
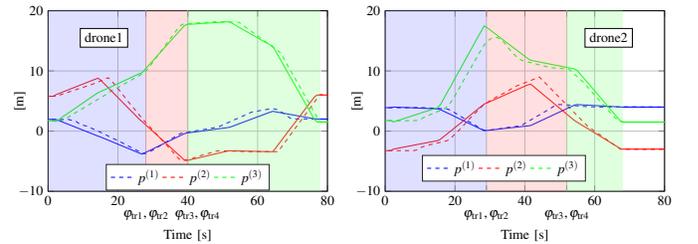

It is worth noticing that the kinodynamic RRT\textsuperscript{$\star$} was not originally meant to 
work with multi-robot systems, making the specification of a minimum distance between drones 
difficult. To overcome this problem, we adapted~\cite{Webb2013ICRA} to the multi-robot case 
via iterating on the various quad-rotors considering the path obtained for the first $\{q-1\}$ 
quad-rotors as forbidden flight corridors for the $q$ quad-rotor. Another important difference is 
that the notion of time is not explicitly taken into account. Besides this, mission 
specifications such as the minimum distance between drones, cannot be easily codified in the 
optimization problem. Furthermore, while the approach in~\cite{Webb2013ICRA} adopts a linearized 
model of the quad-rotor, in our case we do not impose such simplification. From a trajectory 
viewpoint, this implies for~\cite{Webb2013ICRA} spikes and corners difficult to follow in reality, 
possibly leading to sudden stresses on the actuators to respond to rather fast changes of 
direction. This results not only in possibly errors on trajectory tracking, with the possibility of 
violating mission specifications and safety requirements (e.g., the quad-rotor could collide with 
the power tower), but also in high energy consumption that could harm the mission. In addition to 
that, the demand on the onboard control system is higher, since it requires top performances. The 
algorithm took $\SI{8}{\second}$ to find a solution for the problem, while the comparison between 
the drone and desired trajectory from Gazebo simulations is depicted in 
Fig.~\ref{fig:comparisonRRTDroneTrajGazebo}.

%%% END SECTION ============================================================

%%% START SECTION ==========================================================

\section{Experimental Results}
\label{sec:experimentalResults}

%The optimization 
%problem parameter values (see Tab.~\ref{tab:tableParamters}), such as the safety distance 
%$\delta_\mathrm{min}$, maximum velocity $\mathbf{v}^{(j)}_\mathrm{max}$ and acceleration 
%$\mathbf{a}^{(j)}_\mathrm{max}$, were chosen based on the experiment and the hardware setup. 

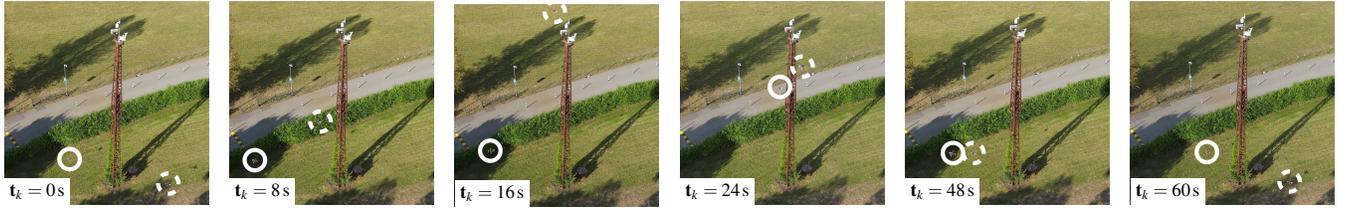
\begin{figure*}
	\begin{center}
		\hspace{-1.42cm}
		\begin{subfigure}{0.155\columnwidth}
		  \centering
		  % left - bottom - right - top
	      \begin{tikzpicture}
	        \node[anchor=south west,inner sep=0] (img) at (0,0) { 
	        \adjincludegraphics[trim={{.30\width} {.15\height} {0.30\width} 
	        {.15\height}},clip,scale=0.150]{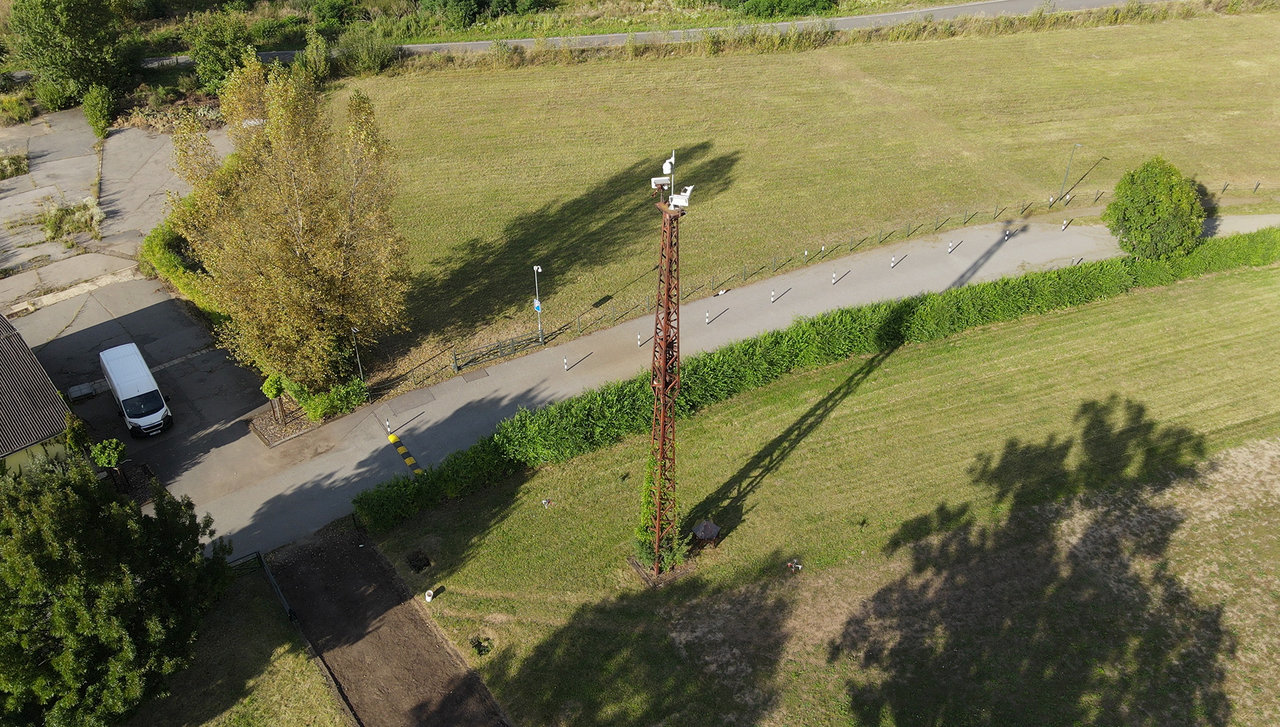}};
	        \begin{scope}[x={(img.south east)},y={(img.north west)}]
	          % plot some stuff over the image
	          \draw [white, dashed, ultra thick] (0.802, 0.105) circle (0.05);
	          \draw [white, ultra thick] (0.32, 0.225) circle (0.05);
	          \node[imglabel,text=black] (label) at (img.south west) {\scriptsize 
	          $\mathbf{t}_k=\SI{0}{\second}$};
	        \end{scope}
	      \end{tikzpicture}
		  \captionsetup[subfigure]{oneside,margin={1cm,1cm}}
		  % \caption{$\mathbf{t}_k=\SI{0}{\second}$}
		\end{subfigure}
		\hspace{1.375cm}
		\begin{subfigure}{0.155\columnwidth}
		  \centering
		  % left - bottom - right - top
	      \begin{tikzpicture}
	        \node[anchor=south west,inner sep=0] (img) at (0,0) { 
	        \adjincludegraphics[trim={{.30\width} {.15\height} {0.30\width} 
	        {.15\height}},clip,scale=0.150]{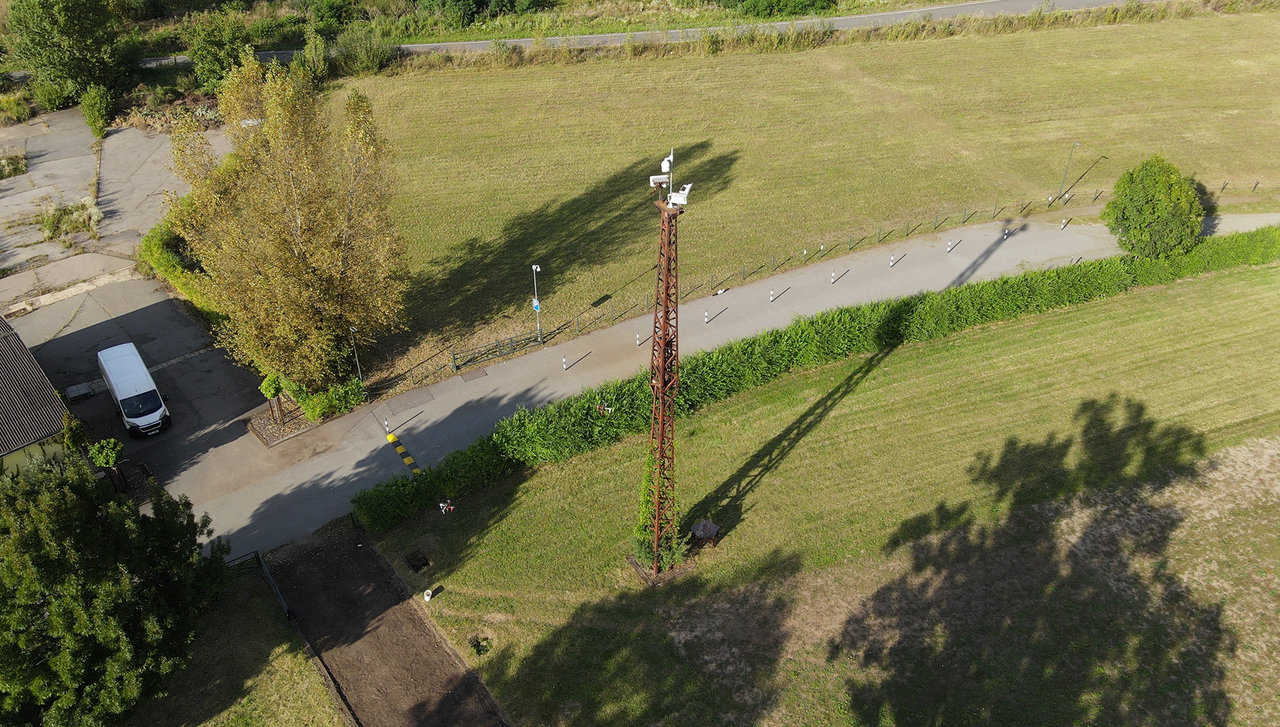}};
	        \begin{scope}[x={(img.south east)},y={(img.north west)}]
	          % plot some stuff over the image
	          \draw [white, ultra thick] (0.125, 0.22) circle (0.05);
	          \draw [white, dashed, ultra thick] (0.445, 0.410) circle (0.05);
	          \node[imglabel,text=black] (label) at (img.south west) {\scriptsize 
	          $\mathbf{t}_k=\SI{8}{\second}$};
	        \end{scope}
	      \end{tikzpicture}
		  % \caption{$\mathbf{t}_k=\SI{12}{\second}$}
		\end{subfigure}
		\hspace{1.375cm}
		\begin{subfigure}{0.155\columnwidth}
		  \centering
		  % left - bottom - right - top
	      \begin{tikzpicture}
	        \node[anchor=south west,inner sep=0] (img) at (0,0) { 
	        \adjincludegraphics[trim={{.30\width} {.15\height} {0.30\width} 
	        {.15\height}},clip,scale=0.150]{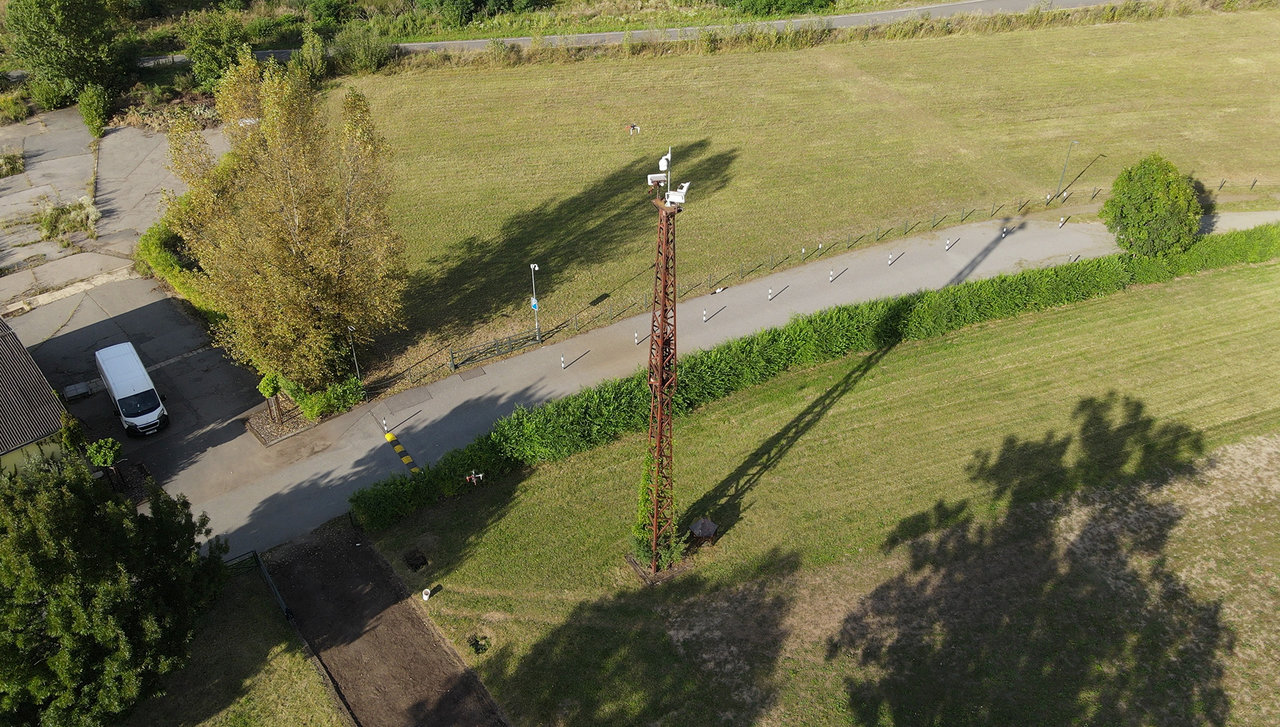}};
	        \begin{scope}[x={(img.south east)},y={(img.north west)}]
	          % plot some stuff over the image
	          \draw [white, ultra thick] (0.175, 0.278) circle (0.05);
	          \draw [white, dashed, ultra thick] (0.490, 0.960) circle (0.05);
	          \node[imglabel,text=black] (label) at (img.south west) {\scriptsize 
	          $\mathbf{t}_k=\SI{16}{\second}$};
	        \end{scope}
	      \end{tikzpicture}
		  % \caption{$\mathbf{t}_k=\SI{24}{\second}$}
		\end{subfigure}
		\hspace{1.375cm}
		\begin{subfigure}{0.155\columnwidth}
		  \centering
		  % left - bottom - right - top
	      \begin{tikzpicture}
	        \node[anchor=south west,inner sep=0] (img) at (0,0) { 
	        \adjincludegraphics[trim={{.30\width} {.15\height} {0.30\width} 
	        {.15\height}},clip,scale=0.150]{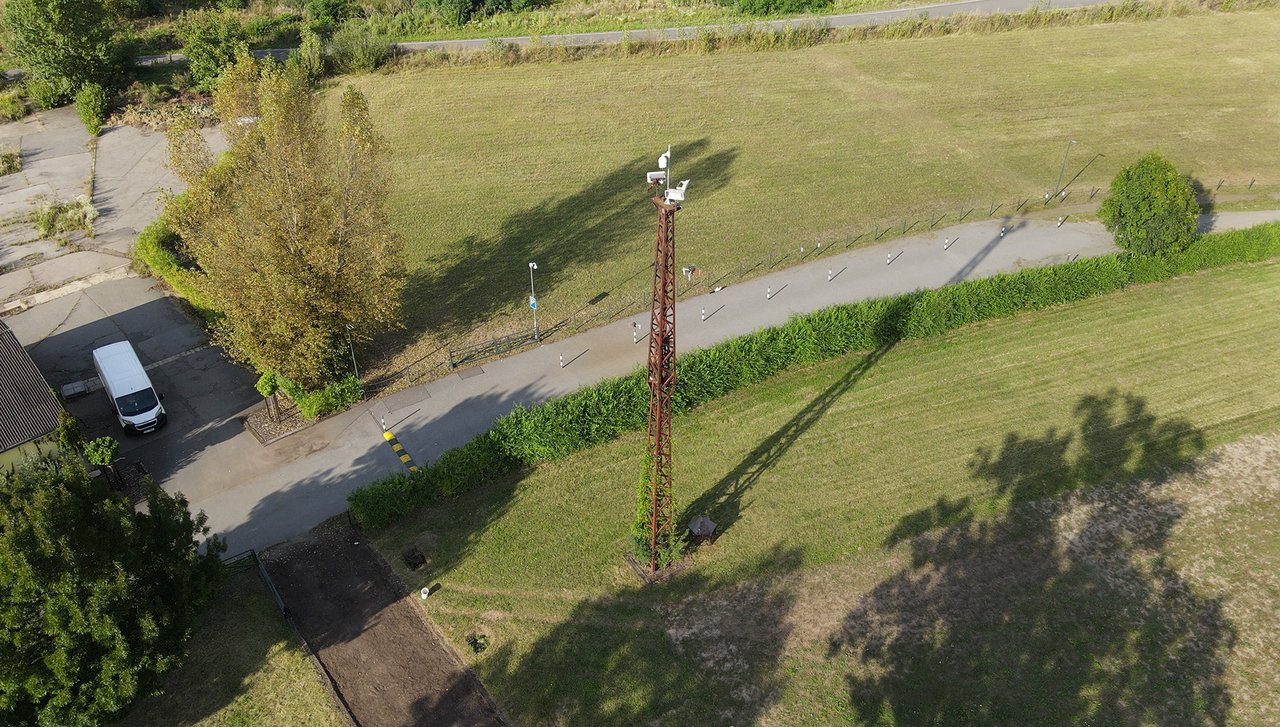}};
	        \begin{scope}[x={(img.south east)},y={(img.north west)}]
	          % plot some stuff over the image
	          \draw [white, dashed, ultra thick] (0.60, 0.68) circle (0.05);
	          \draw [white, ultra thick] (0.49, 0.58) circle (0.05);
	          \node[imglabel,text=black] (label) at (img.south west) {\scriptsize 
	          $\mathbf{t}_k=\SI{24}{\second}$};
	        \end{scope}
	      \end{tikzpicture}
		  % \caption{$\mathbf{t}_k=\SI{36}{\second}$}
		\end{subfigure}
		\hspace{1.375cm}
		\begin{subfigure}{0.155\columnwidth}
		  \centering
		  % left - bottom - right - top
	      \begin{tikzpicture}
	        \node[anchor=south west,inner sep=0] (img) at (0,0) { 
	        \adjincludegraphics[trim={{.30\width} {.15\height} {0.30\width} 
	        {.15\height}},clip,scale=0.150]{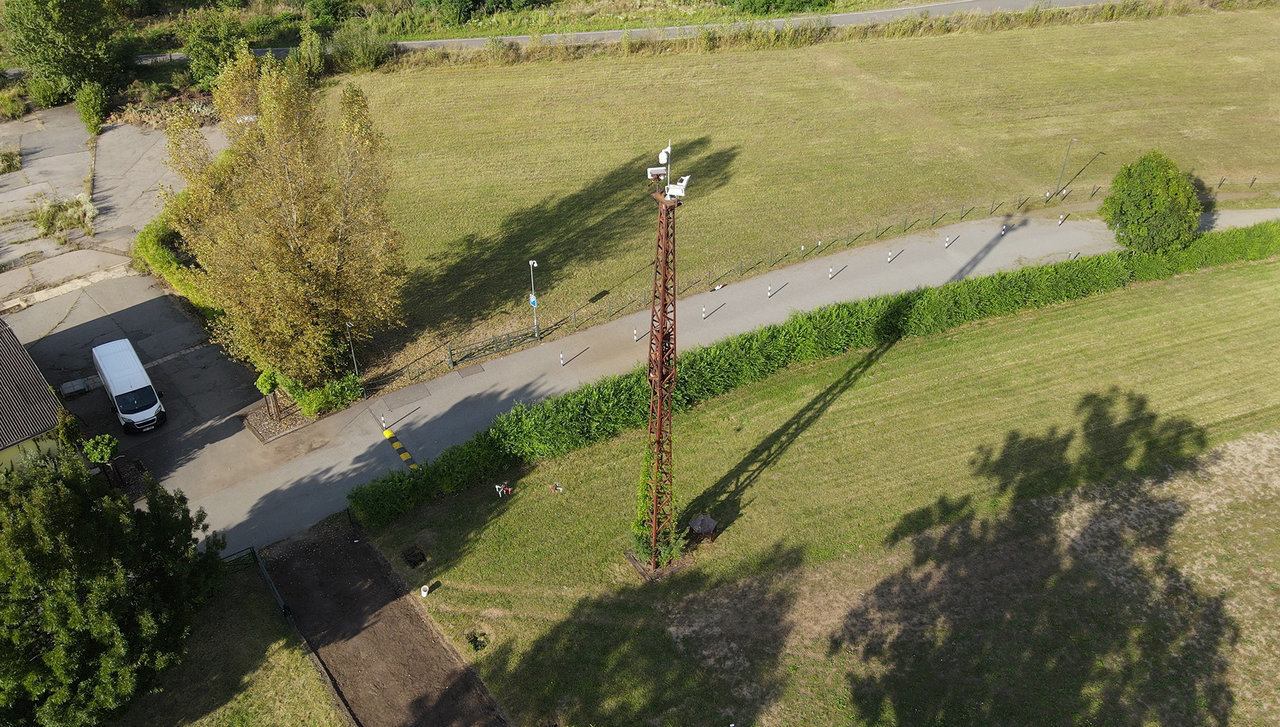}};
	        \begin{scope}[x={(img.south east)},y={(img.north west)}]
	          % plot some stuff over the image
	          \draw [white, dashed, ultra thick] (0.335, 0.255) circle (0.05);
	          \draw [white, ultra thick] (0.235, 0.255) circle (0.05);
	          \node[imglabel,text=black] (label) at (img.south west) {\scriptsize 
	          $\mathbf{t}_k=\SI{48}{\second}$};
	        \end{scope}
	      \end{tikzpicture}
		  % \caption{$\mathbf{t}_k=\SI{48}{\second}$}
		\end{subfigure}
		\hspace{1.375cm}
		\begin{subfigure}{0.155\columnwidth}
		  \centering
		  % left - bottom - right - top
	      \begin{tikzpicture}
	        \node[anchor=south west,inner sep=0] (img) at (0,0) { 
	        \adjincludegraphics[trim={{.30\width} {.15\height} {0.30\width} 
	        {.15\height}},clip,scale=0.150]{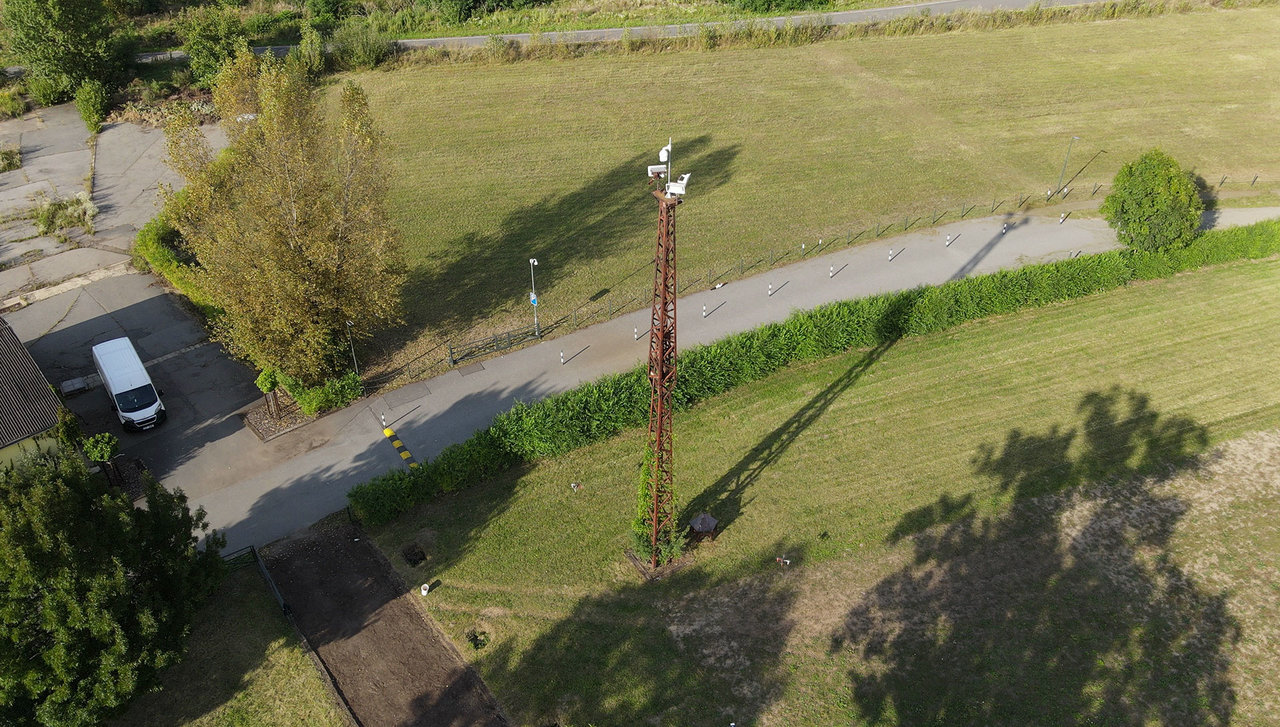}};
	        \begin{scope}[x={(img.south east)},y={(img.north west)}]
	          % plot some stuff over the image
	          \draw [white, dashed, ultra thick] (0.779, 0.115) circle (0.05);
	          \draw [white, ultra thick] (0.375, 0.260) circle (0.05);
	          \node[imglabel,text=black] (label) at (img.south west) {\scriptsize 
	          $\mathbf{t}_k=\SI{60}{\second}$};
	        \end{scope}
	      \end{tikzpicture}
		  % \caption{$\mathbf{t}_k=\SI{60}{\second}$}
		\end{subfigure}
		%
		%\hspace{1.25cm}
%		\vspace{0.25cm}
%		\\
		%%%%%%%%%%%%%%%%%%%%%%%%%%% MULTI-LINE
%		\hspace{-3.75cm}
%		%
%		\begin{subfigure}{0.20\columnwidth}
%			\centering
%			% left - bottom - right - top
%			\adjustbox{trim={.05\width} {.25\height} {0.05\width} {.25\height},clip}%
%			{\includegraphics[scale=0.1, height=4.5cm,
%			width=6cm]{figure/experiment_photos_longRange_selected/scene00901-1.jpg}}
%			\caption{$\mathbf{t}_k=\SI{0}{\second}$}
%		\end{subfigure}
		%
%		\hspace{3.75cm}
		%
%		\begin{subfigure}{0.20\columnwidth}
%			\centering
%			% left - bottom - right - top
%			\adjustbox{trim={.0\width} {.25\height} {0.0\width} {.25\height},clip}%
%			
%{\includegraphics[scale=0.1]{figure/experiment_photos_longRange_selected/scene01501-1.jpg}}
%			\caption{$\mathbf{t}_k=\SI{0}{\second}$}
%		\end{subfigure}
		%
%		\begin{subfigure}{0.20\columnwidth}
%			\centering
%			% left - bottom - right - top
%			\adjustbox{trim={.05\width} {.25\height} {0.05\width} {.25\height},clip}%
%{\includegraphics[scale=0.1, height=4.5cm,
%width=6cm]{figure/experiment_photos_longRange_selected/scene01651-1.jpg}}
%			\caption{$\mathbf{t}_k=\SI{0}{\second}$}
%		\end{subfigure}
		%
%		\hspace{3.75cm}
		%
%		\begin{subfigure}{0.20\columnwidth}
%			\centering
%			% left - bottom - right - top
%			\adjustbox{trim={.05\width} {.25\height} {0.05\width} {.25\height},clip}%			
%{\includegraphics[scale=0.1, height=4.5cm,
%width=6cm]{figure/experiment_photos_longRange_selected/scene02101-1.jpg}}
%			\caption{$\mathbf{t}_k=\SI{0}{\second}$}
%		\end{subfigure}
	\end{center}
	\vspace{-2mm}
	\caption{Snapshots of the \textit{power tower inspection} scenario. The system evolution at 
	different time instants $\mathbf{t}_k$ is reported. Solid and dashed circles are used to 
	indicate ``drone1'' and ``drone2'', respectively.}
	\label{fig:frameRealExperiments}
  \vspace{-1em}
\end{figure*}

To evaluate and prove the applicability of the proposed approach in real-world autonomous 
inspection tasks, experiments with a DJI F450 quad-rotor were performed (see 
Fig.~\ref{fig:frameRealExperiments}). Real flight tests verified not only the fulfillment of the 
STL specifications (i.e., $\varphi_\mathrm{tr1}$, $\varphi_\mathrm{tr2}$, $\varphi_\mathrm{tr3}$, 
and $\varphi_\mathrm{tr4}$), but also the compliance with trajectory generation requirements (i.e., 
maximum velocity $\mathbf{v}^{(j)}_\mathrm{max}$ and acceleration $\mathbf{a}^{(j)}_\mathrm{max}$, 
and safe distance $\delta_\mathrm{min}$). The STL motion planner (see 
Sec.~\ref{sec:problemFormulation}) was implemented in MATLAB and the obtained trajectories were 
sent to onboard PCs before running the experiment. 

%The maximum velocity value of the agents should be chosen in accordance with the camera 
%constraints, otherwise it would result in blurred images. Additionally, the velocity of the 
%quad-rotors should be greater than the wind gust in order to compensate it and, for safety 
%reasons, the velocity should also be reasonable for the safety pilot to be able to take control of 
%the vehicle in the event of a malfunction. The safety distance should not be less than the GPS 
%accuracy, in order to avoid drift and environmental factors that may impact system location, 
%leading the drones to possible crashes.

%%% END SECTION ============================================================

%%% START SECTION ==========================================================

\section{Conclusions}
\label{sec:conclusions}

This paper has presented a framework for encoding power line inspection missions for a fleet of 
quad-rotors as STL specifications. In particular, an optimization problem was set to generate 
optimal strategies that satisfy such specifications accounting also for vehicle constraints. 
Further, an event-triggered replanner and an energy minimization problem have been proposed 
to reply to external disturbances and to enhance quad-rotors energy saving during the mission. The 
approach enables the use of complex and rich task specifications with automated trajectory 
generation. The solution potentially mitigates computational complexity explosion issues, thanks to 
the use of a smooth approximation of the robust semantic. The numerical simulations in MATLAB and 
Gazebo and also the experimental results proved the validity and the effectiveness of the proposed 
approach, demonstrating its applicability in real world scenarios. Future work includes 
investigating better solutions to cluster target regions accounting for drone positions and 
the distance between targets and extending the event-triggered replanner with an online search of 
the best partial reconnection trajectory or, if complete specification satisfaction is no longer 
possible, the minimum violation replanning.

%%% END SECTION ============================================================

%%% START SECTION ==========================================================
	
\vspace{-3.5mm}
\bibliographystyle{IEEEtran}
\bibliography{bib_short.bib}

\begin{thebibliography}{10}
\providecommand{\url}[1]{#1}
\csname url@samestyle\endcsname
\providecommand{\newblock}{\relax}
\providecommand{\bibinfo}[2]{#2}
\providecommand{\BIBentrySTDinterwordspacing}{\spaceskip=0pt\relax}
\providecommand{\BIBentryALTinterwordstretchfactor}{4}
\providecommand{\BIBentryALTinterwordspacing}{\spaceskip=\fontdimen2\font plus
\BIBentryALTinterwordstretchfactor\fontdimen3\font minus
  \fontdimen4\font\relax}
\providecommand{\BIBforeignlanguage}[2]{{%
\expandafter\ifx\csname l@#1\endcsname\relax
\typeout{** WARNING: IEEEtran.bst: No hyphenation pattern has been}%
\typeout{** loaded for the language `#1'. Using the pattern for}%
\typeout{** the default language instead.}%
\else
\language=\csname l@#1\endcsname
\fi
#2}}
\providecommand{\BIBdecl}{\relax}
\BIBdecl

\bibitem{EPRI2011TechnicalReport}
J.~{Major} \emph{et~al.}, ``{Emerging and future inspection of overhead
  transmission lines},'' EPR Institute, Tech. Rep., 2011, no. 1021876.

\bibitem{Baik2018JIRS}
H.~{Baik} and J.~{Valenzuela}, ``{Unmanned Aircraft System Path Planning for
  Visually Inspecting Electric Transmission Towers},'' \emph{Journal of
  Intelligent \& Robotic Systems}, vol.~95, pp. 1097--1111, 2018.

\bibitem{Martinez2018EAAI}
C.~{Martinez} \emph{et~al.}, ``{The Power Line Inspection Software (PoLIS): A
  versatile system for automating power line inspection},'' \emph{Engineering
  Applications of Artificial Intelligence}, vol.~71, pp. 293--314, 2018.

\bibitem{Donze2010FMATS}
A.~{Donz{\'e}} and O.~{Maler}, ``{Robust Satisfaction of Temporal Logic over
  Real-Valued Signals},'' in \emph{Formal Modeling and Analysis of Timed
  Systems}, K.~Chatterjee and T.~A. Henzinger, Eds., 2010, pp. 92--106.

\bibitem{Maler2004FTMA}
O.~{Maler} and D.~{Nickovic}, ``{Monitoring Temporal Properties of Continuous
  Signals},'' in \emph{{Formal Techniques, Modelling and Analysis of Timed and
  Fault-Tolerant Systems}}, 2004, pp. 152--166.

\bibitem{Webb2013ICRA}
D.~J. {Webb} \emph{et~al.}, ``{Kinodynamic RRT*: Asymptotically optimal motion
  planning for robots with linear dynamics},'' in \emph{IEEE International
  Conference on Robotics and Automation}, 2013, pp. 5054--5061.

\bibitem{Belta2017Book}
C.~{Belta}, B.~{Yordanov}, and E.~A. {Gol}, \emph{{Formal Methods for
  Discrete-time Dynamical Systems}}.\hskip 1em plus 0.5em minus 0.4em\relax
  Springer, 2017.

\bibitem{Faunekos2009TCS}
G.~E. {Fainekos} and G.~J. {Pappas}, ``{Robustness of temporal logic
  specifications for continuous-time signals},'' \emph{Theoretical Computer
  Science}, vol. 410, no.~42, pp. 4262--4291, 2009.

\bibitem{Pagnano2013CIRP}
A.~{Pagnano}, M.~{H\"{o}pf}, and R.~{Teti}, ``{A Roadmap for Automated Power
  Line Inspection. Maintenance and Repair},'' \emph{{Procedia CIRP}}, vol.~12,
  pp. 234--239, 2013.

\bibitem{Chen2019Access}
H.~{Chen}, Z.~{He}, B.~{Shi} \emph{et~al.}, ``{Research on Recognition Method
  of Electrical Components Based on YOLO V3},'' \emph{IEEE Access}, vol.~7, pp.
  157\,818--157\,829, 2019.

\bibitem{Mansouri2018CEP}
S.~S. {Mansouri}, C.~{Kanellakis}, E.~{Fresk} \emph{et~al.}, ``{Cooperative
  coverage path planning for visual inspection},'' \emph{Control Engineering
  Practice}, vol.~74, pp. 118--131, 2018.

\bibitem{Shoukry2017CDC}
Y.~{Shoukry}, P.~{Nuzzo}, A.~{Balkan} \emph{et~al.}, ``{Linear temporal logic
  motion planning for teams of underactuated robots using satisfiability modulo
  convex programming},'' in \emph{IEEE Conference on Decision and Control},
  2017, pp. 1132--1137.

\bibitem{Luis2020RAL}
C.~E. {Luis} \emph{et~al.}, ``{Online Trajectory Generation With Distributed
  Model Predictive Control for Multi-Robot Motion Planning},'' \emph{IEEE
  Robotics and Automation Letters}, vol.~5, no.~2, pp. 604--611, 2020.

\bibitem{Honig2018TRO}
W.~{H\"{o}nig},  \emph{et~al.}, ``{Trajectory Planning for Quadrotor Swarms},''
  \emph{IEEE Transactions on Robotics}, vol.~34, no.~4, pp. 856--869, 2018.

\bibitem{PantCCT2017}
Y.~V. {Pant} \emph{et~al.}, ``{Smooth operator: Control using the smooth
  robustness of temporal logic},'' in \emph{IEEE Conference on Control
  Technology and Applications}, 2017, pp. 1235--1240.

\bibitem{Park2019IROS}
J.~{Park} \emph{et~al.}, ``{Fast Trajectory Planning for Multiple Quadrotors
  using Relative Safe Flight Corridor},'' in \emph{IEEE International
  Conference on Intelligent Robots and Systems}, 2019, pp. 596--603.

\bibitem{Raman2014CDC}
V.~{Raman}, A.~{Donzé} \emph{et~al.}, ``{Model predictive control with signal
  temporal logic specifications},'' in \emph{IEEE Conference on Decision and
  Control}, 2014, pp. 81--87.

\bibitem{Mueller2015TRO}
M.~W. {Mueller}, M.~{Hehn}, and R.~{D'Andrea}, ``{A Computationally Efficient
  Motion Primitive for Quadrocopter Trajectory Generation},'' \emph{IEEE
  Transactions on Robotics}, vol.~31, no.~6, pp. 1294--1310, 2015.

\bibitem{Baca2020mrs}
\BIBentryALTinterwordspacing
T.~{Baca}, M.~{Petrlik}, M.~{Vrba} \emph{et~al.}, ``{The MRS UAV System:
  Pushing the Frontiers of Reproducible Research, Real-world Deployment, and
  Education with Autonomous Unmanned Aerial Vehicles},'' 2020. [Online].
  Available: \url{https://arxiv.org/pdf/2008.08050}
\BIBentrySTDinterwordspacing

\bibitem{Abbas2013ACM}
H.~{Abbas}, G.~{Fainekos}, S.~{Sankaranarayanan} \emph{et~al.},
  ``{Probabilistic Temporal Logic Falsification of Cyber-Physical Systems},''
  \emph{ACM Transaction on Embedded Computing Systems}, vol.~12, no.~2s, 2013.

\bibitem{Silano2019SMC}
G.~{Silano} \emph{et~al.}, ``{Software-in-the-loop simulation for improving
  flight control system design: a quadrotor case study},'' in \emph{IEEE
  International Conference on Systems, Man, and Cybernetics}, 2019, pp.
  466--471.

\end{thebibliography}

\end{document}